\documentclass{article} 
\usepackage[final]{colm2026_conference}

\usepackage{microtype}
\usepackage{hyperref}
\usepackage{url}
\usepackage{booktabs}
\usepackage{graphicx}

\usepackage{lineno}
\usepackage{xspace}
\usepackage{amsmath}
\usepackage[capitalise]{cleveref}
\usepackage{subcaption}
\usepackage{multirow}
\usepackage{wrapfig}
\usepackage{makecell}
\usepackage{enumitem}

\definecolor{darkblue}{rgb}{0, 0, 0.5}
\hypersetup{colorlinks=true, citecolor=darkblue, linkcolor=darkblue, urlcolor=darkblue}

\title{Data-Efficient Adaptation of LLMs via  \\ Attention Head Reweighting}

\author{Tuomas Oikarinen \\ UC San Diego, Microsoft Research
\And
Zixiao Chen\\
Microsoft Security AI\\
\And
Charlotte Siska \\
Microsoft Security AI\\
\AND
Tsui-Wei Weng \\
UC San Diego
\And
Chandan Singh\\
Microsoft Research
\And
Jianfeng Gao \\
Microsoft Research
}

\newcommand{\methodlong}{Attention Head Reweighting\xspace}
\newcommand{\method}{AHR\xspace}

\begin{document}

\ifcolmsubmission
\linenumbers
\fi

\maketitle

\begin{abstract}
Learning effectively from limited data is critical in domains like security where labeled examples are scarce.
Large language models (LLMs) have demonstrated some capabilities for data-efficient learning, especially through parameter-efficient adaptation methods, but 
continue to struggle when faced with few samples for difficult tasks.
To meet this challenge, we propose \methodlong (\method), a data-efficient method that adapts LLMs to new text-classification tasks by learning only a single scalar per attention head.
This drastically reduces the number of parameters that need to be learned by making use of the functional specialization of individual attention heads.
Experiments on diverse open-source text classification datasets show that \method can outperform standard baselines like LoRA when learning from limited samples, despite having $200-1000\times$ fewer trainable parameters, as our AHR only modifies $\sim 0.0001\%$ of the model's parameters. In addition, our learned weights are easy to interpret and can be analyzed to better understand the mechanisms and attention heads responsible for in-context learning abilities in LLMs.\footnote{Code to use \method (fully compatible with the \href{https://github.com/huggingface/peft}{peft package}) and to reproduce our experiments will be available on Github at \href{https://github.com/tuomaso/attention-head-reweighting}{github.com/tuomaso/attention-head-reweighting}}
\end{abstract}

\section{Introduction}

Learning effectively from limited labeled data remains a central challenge in high-stakes domains
such as AI security, where annotations are often scarce because attacks and threat vectors are constantly evolving and need to be defended against quickly~\citep{divakaran2024llms}.
In this setting, data-efficient text classification is critical for flagging large language model (LLM) inputs~\citep{yi2024jailbreak,saha2024llm,chao2025jailbreaking,verma2025multiguard} and LLM outputs~\citep{inan2023llama} that can be dangerous.

While LLMs exhibit some data-efficient behavior through in-context learning (ICL)~\citep{brown2020language,openai2023gpt4} and parameter-efficient fine-tuning (PEFT)~\citep{hu2022lora,zhang2023adalora}, their performance often degrades sharply when only a small number of task-specific examples are available.
One issue with these methods is that they generally optimize in a large, unconstrained parameter space, which can lead to overfitting.

\begin{figure}[t]
    \centering
        \includegraphics[width=0.85\linewidth]{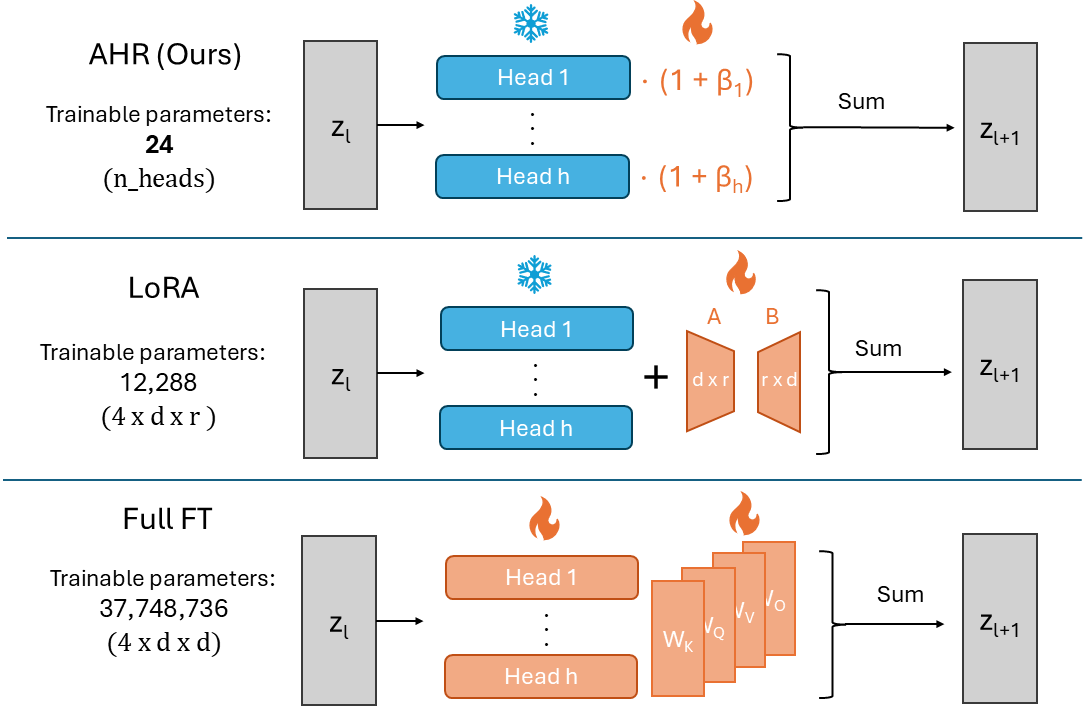}
    \caption{\textit{\method overview}. Simplified visual comparison between our \method, LoRA and full-finetuning on a single attention layer. Parameter counts are calculated for a standard transformer with $d=3072$, with $r=1$ for LoRA. We can see AHR reduces trainable parameters $~500 \times$ compared to LoRA.}
    \label{fig:overview}
\end{figure}

To tackle this issue, we propose Attention Head Reweighting (\method), an intuitive data-efficient method to adapt LLMs to new text-classification tasks by learning only a single scalar per attention head.
This drastically reduces the number of parameters that need to be adapted (see \cref{fig:overview}).
We hypothesize our method is effective and avoids overfitting by leveraging the functional specialization of attention heads noted in recent works~\citep{olsson2022context,akyurek2024context,zhang2023tell,ge2023model}.
For example, if a particular security-related behavior is isolated to a few attention heads, \method can easily learn to change the weighting for these heads rather than seeking to alter the internal weights of each head.

In our experiments we focus on text classification tasks, as these can often be effectively learned from a few examples, and many important security tasks such as detecting phishing attempts or model jailbreaks are text classification tasks. Classifiers are a critical bottleneck in AI security, which is constantly faced with data scarcity as new threats emerge, for filtering and guarding LLM responses that are interacting with real-world users and attackers.

In our experiments, \method outperforms baselines across a variety of settings and models, improving by an average of around $3\%$ points over standard baselines like LoRA \citep{hu2022lora} when training data is very limited i.e. $
\leq 100$ examples.
\method shows particularly large gains in the security relevant tasks of Webpage Phishing Detection (Web) and Jailbreak detection, achieving 6-7\% accuracy improvement over best baseline methods with just 10 training samples, while using $200-1000 \times$ fewer trainable parameters.
We can analyze the changes our method makes to the model, and see that these gains are largely fueled by a combination of upweighting a few task specific heads as well as a couple general in-context learning heads.
Finally, to maximally learn from just a few examples, we utilize in-context finetuning (IC-FT), where we update model parameters to maximize the probability of correct answers within a few-shot prompt, which we show improves accuracy by $\sim 10\%$ points over finetuning on just one sample at a time.

\section{Background and related work}

\paragraph{Parameter-efficient finetuning}
A variety of parameter-efficient finetuning methods have been proposed~ to adapt large models while updating only a small subset of parameters~\citep{zhang2025parameterefficientfinetuningfoundationmodels,han2024parameterlarge,wang202410parameter}.
Prominent examples include LoRA~\citep{hu2022lora}, AdaLoRA~\citep{zhang2023adalora}, (IA)\textsuperscript{3}~\citep{liu2022few}, 
and related approaches~\citep{hayou2024lora+,hayou2025plop,logan2022cutting}.
These methods substantially reduce adaptation cost, but still introduce thousands to millions of trainable parameters, which can lead to overfitting in  data-scarce settings.

Simultaneously, there has been a spectrum of increasingly constrained prompting approaches, ranging from continuous prompt tuning~\citep{Li2021PrefixTuningOC,Liu2021PTuningVP}
to discrete prompt tuning~\citep{shin2020autoprompt}, 
prompt ensemble construction~\citep{hou2022promptboosting,pitis2023boosted,morris2023tree},
and generating a single natural-language prompt~\citep{zhou2022large,singh2023explaining}.
While effective in some settings, prompt-based methods are often very sensitive to minor variations and may struggle with under-specified tasks under severe data scarcity.

\paragraph{In-context learning}
ICL obviates the need for task-specific parameter adaptation altogether, but can be highly sensitive to minor variations in the provided examples~\citep{min2022rethinking} and sometimes unfaithful to those examples~\citep{wei2023larger,webson2022prompt}.
ICL can be improved through directed instruction tuning~\citep{flant5} or by including explanations along with examples~\citep{lampinen2022can}.
A growing body of work has studied ICL, showing that it can implement linear models~\citep{akyurek2022learning,zhang2023trained}, discrete functions~\citep{bhattamishra2023understanding}, and more general algorithms~\citep{li2023transformers,zhuang2025vector}.
Some works have further argued that ICL implicitly performs optimization steps analogous to gradient descent~\citep{mahankali2023one,von2023transformers,ahn2024transformers} and higher-order optimization methods~\citep{dai2023can,zhang2023trained}.

\begin{table*}[]
\centering
\begin{tabular}{@{}lrrrr@{}}
\toprule
Model & \multicolumn{1}{l}{GPT2-XL} & \multicolumn{1}{l}{Llama-3.2-1B} & \multicolumn{1}{l}{Llama-3.2-3B} & \multicolumn{1}{l}{QWEN3-8B} \\ \midrule
Full FT & 1,558,148,000 & 1,235,920,896 & 3,212,750,496 & 8,190,736,512 \\
LoRA & 307,200 & 106,496 & 286,720 & 479,232 \\
AdaLoRA & 614,496 & 213,056 & 573,552 & 958,608 \\
IA3 & 537,600 & 147,456 & 286,720 & 626,688 \\
AHR (Ours) & \textbf{1,200} & \textbf{512} & \textbf{672} & \textbf{1,152} \\ \bottomrule
\\
\end{tabular}
\caption{Comparing the number of trainable parameters for different models and different parameter efficient finetuning methods. Our \method requires $200-1,000\times$ less trainable parameters than the baseline methods. In total we train less than one millionth of the model's parameters.}
\label{tab:trainable_params}
\end{table*}

\paragraph{Attention heads}
Attention heads have been a primary focus for analysis~\citep{zheng2024attention,bibal2022attention}, with evidence that attention heads often latch onto task-specific patterns that are useful for ICL, e.g. induction heads~\citep{olsson2022context}, n-gram heads~\citep{akyurek2024context}, and more easily explainable patterns~\citep{todd2025context,oikarinen2024linear,singh2023explainingmodules}.
This growing evidence of functional specialization has motivated methods that explicitly manipulate attention.
Prior works have shown that modifying attention scores can be useful for post-hoc steering~\citep{zhang2023tell,zhang2024model}, 
attention steering for transfer learning~\citep{shi2023toast},
attention-guided retrieval~\citep{zhang2025attention},
interpretable modeling~\citep{kim2024interpretable},
boosting attention for instruction following~\citep{guardieiro2025instruction},
and improving long-context LLM inference~\citep{gu2024llmsteer}.

\section{Method: \methodlong}

Our paper aims to improve the efficiency of how models can learn from a very small number of training examples by modifying an extremely small number of parameters. We do this via \textbf{Attention Head Reweighting (AHR)}, which introduces a single learnable scalar parameter per attention head that multiplies that head's contributions to the residual stream (see \cref{fig:overview}).
By finetuning these parameters on a small amount of data, we can amplify attention heads that are helpful for the current task, while downweighting heads that harm performance. By not modifying the weights inside individual heads we can avoid overfitting to the small amount of training data. 

\textbf{Notation:} We follow the notation of \citet{elhage2021mathematical}, where we divide the $W_O$ matrix into its components within each attention head. Let $z_r^l \in \mathbb{R}^{s \times d}$ be the model's residual stream after layer $l$, where $s$ is the number of tokens in the sequence and $d$ is the model's representation dimension. A typical transformer block consists of a Multihead-Attention (MHA) layer followed by a multi-layer perceptron (MLP) layer. We can write the update of a layer as:
\begin{equation}
    z^{l+1}_{Att} = z_r^l + \sum_{h \in H^{l+1}}h(z^l_r)
    \label{eq:orig_attn}
\end{equation}
\begin{equation}
    z_r^{l+1} = z_{Att}^{l+1} + MLP^{l+1}(z_{Att}^{l+1})
\end{equation}

Here $h(z) = AzW_VW_O$, $A = \text{softmax}(\frac{(zW_Q)(zW_K)^T}{\sqrt{d_k}})$. $W_V \in \mathbb{R}^{d \times d_k}$, $W_Q \in \mathbb{R}^{d \times d_k}$, $W_K \in \mathbb{R}^{d \times d_k}$ and $W_O \in \mathbb{R}^{d_k \times d}$ are the parameters of the attention head, and $d_k = d/n_h$ where $n_h = |\mathcal{H}|$ is the number of attention heads per layer.

\paragraph{Attention Head Reweighting:} Our method simply edits the update rule of the multi-head attention layer (Eq. \ref{eq:orig_attn}) to:

\begin{equation}
    z^{l+1}_{Att} = z_r^l + \sum_{h \in H^{l+1}}(1 + \beta_h) h(z^l_r),
    \label{eq:modified_attn}
\end{equation}

where $\beta_h$ is a head-specific parameter that is initialized at zero. During training, we only train the $\beta_h$ parameters while keeping all other parameters, including the MLP layers unchanged. At test time, the parameters $\beta$ can be merged into $W_O$ for no additional inference cost. During training, we utilize supervised fine-tuning (SFT) on few-shot examples, by minimizing the cross-entropy loss on the answer tokens, as well as a regularization objective to minimize changes to the model:

\begin{equation}
    \min_{\beta} \sum_{(x, y) \in \mathcal{D}_{train}} -\log P(y |x; \theta, \beta) + \lambda (\sum_{h} |\beta_h|^p)^{1/p}
\end{equation}

where $\mathcal{D}_{train}$ is the few-shot training dataset, $\lambda$ is a hyperparameter controlling regularization and $\theta$ are the (frozen) model parameters and $p \in \{1, 2\}$.

\subsection{In-Context Finetuning}
\label{sec:ic_ft}

Recently, In-Context Learning (ICL) has risen as an alternative to finetuning as a way for models to learn from few-shot examples \citep{brown2020language}. While powerful, in-context learning is limited in how many samples it can learn from by the model's context window, as well as GPU-memory. Additionally, using many in-context samples incurs a large additional cost at inference time. 

To get the best of both worlds, we utilize In-Context finetuning (IC-FT) similar to \citep{he2025finetuneoncemanyshotincontext}. Different from \citep{he2025finetuneoncemanyshotincontext}, we finetune separately on each dataset while their goal is to improve general in-context learning ability. In in-context finetuning, we construct an input string with $k$ few-shot examples, and do supervised finetuning to maximize the probability of all answer tokens within the context. For each input with $k$ in-context examples, we minimize the following:

\begin{equation}
    \min_{\beta} \sum_{i=1}^k -\log P(y_i | X_i; \theta, \beta) + \lambda (\sum_{h} |\beta_h|^p)^{1/p}
    \label{eq:ic_ft}
\end{equation}

where $X_i = \{x_1, y_1, ..., y_{i-1}, x_i\}$ is the concatenated input with few-shot examples. As this technique is independent of the finetuning method, we utilize it for both our method and the baselines in our experiments (with their trainable parameters instead of $\beta$ in Eq. \ref{eq:ic_ft}). For text classification, we only maximize the probability of the first token in the answer, as predicting subsequent tokens of the answer is trivial given the first one.

\begin{figure}
    \centering
    \begin{subfigure}[b]{0.48\linewidth}
        \centering
        \includegraphics[width=\linewidth]{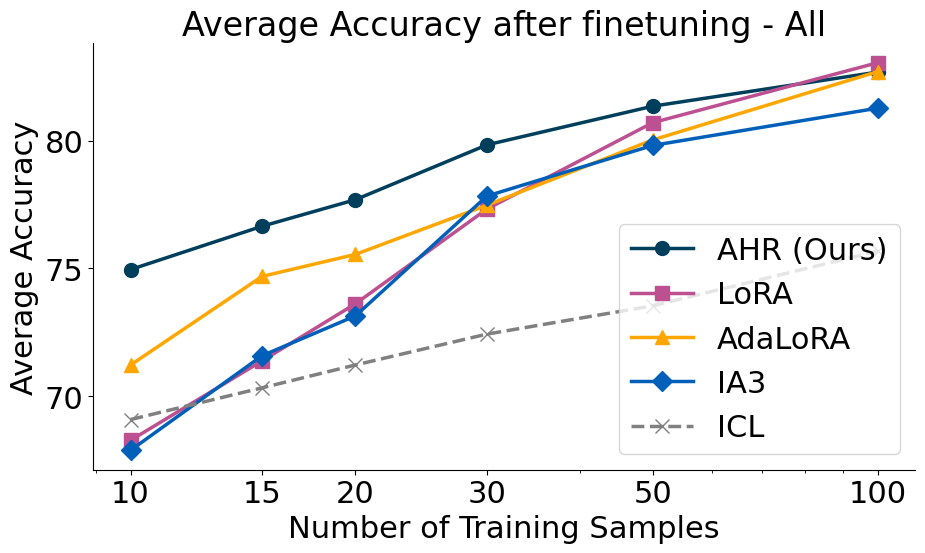}
        \caption{Averaged across 6 datasets}
        \label{fig:classif_all}
    \end{subfigure}
    \hfill
    \begin{subfigure}[b]{0.48\linewidth}
        \centering
        \includegraphics[width=\linewidth]{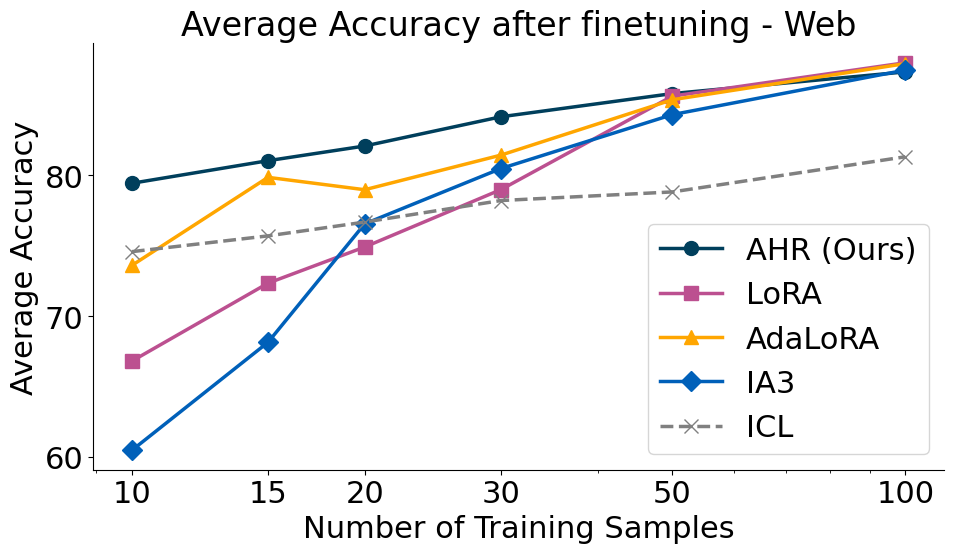}
        \caption{\textit{Web} - Phishing URL classification}
        \label{fig:classif_web}
    \end{subfigure}
    \caption{Accuracy after finetuning, averaged across our 4 models. We can see our AHR significantly outperforms baselines when data is limited, while performance starts to equalize at around 100 samples.}
    \label{fig:classif}
\end{figure}

\section{Results}

\subsection{Experimental setup}

\paragraph{Models}
We consider four LLMs across a range of sizes in our experiments: 
GPT-2 XL~(\citeauthor{radford2019language}~\citeyear{radford2019language}; \texttt{gpt2-xl}),
LLaMa-3.2 1B and 3B~(\citeauthor{dubey2024llama}~\citeyear{dubey2024llama}; \texttt{meta-llama/Llama-3.2-1B} and \texttt{meta-llama/Llama-3.2-3B}),
and Qwen3 8B (\citeauthor{yang2025qwen3}~\citeyear{yang2025qwen3}; \texttt{Qwen/Qwen3-8B}). See Table \ref{tab:trainable_params} for more details, such as the number of parameters in each model and how many of these are trainable.

\paragraph{Datasets}
We test our method on a set of six text-classification datasets, including standard tasks as well as cybersecurity relevant applications:
\begin{itemize}[noitemsep,topsep=0pt,leftmargin=*]
    \item SST2~\citep{socher2013recursive}. This dataset contains sentences from movie reviews, and the task is to classify their sentiment into one of \{Positive, Negative\}.
    \item AG-News~\citep{zhang2015character}. A collection of text excerpts from news articles, where the goal is to classify the excerpt into one of the following categories: \{World, Sports, Business, Sci/Tech\}
    \item Emotion~\citep{saravia2018carer}. A dataset of tweets, with labels based on the emotion expressed in the tweet. This dataset contains 6 classes \{Sadness, Joy, Love, Anger, Fear, Surprise\}
    \item Webpage Phishing Detection (Web)~\citep{Hannousse2021WebPhishing}, a dataset of web URLs where the goal is to differentiate between Phishing and Legitimate URLs.
    \item Toxigen~\citep{hartvigsen2022toxigen}. This is a dataset of potentially toxic/hateful comments for detecting hateful speech. We treated an input as toxic if the average human assessed toxicity score was $>3$ (on a 1-5 scale) and as benign otherwise.
    \item Jailbreak Detection~\citep{shen2024characterizing}, a dataset of LLM prompts containing Jailbreak attempts as well as standard prompts. The goal is to classify whether a given prompt is a jailbreak attempt. We use the balanced version from \href{https://huggingface.co/datasets/jackhhao/jailbreak-classification}{https://huggingface.co/datasets/jackhhao/jailbreak-classification}.
    
\end{itemize}

Most of these datasets have thousands of examples, but in order to study few-shot learning capabilities of our models, we randomly sample a subset of the training data to be used in few-shot learning. We evaluate performance on the full test set. See an example for each dataset in \cref{tab:data_examples}.


\paragraph{Baselines:} As our main baseline, we compare against LoRA \citep{hu2022lora}. For LoRA we use $r=1$ and $\alpha=1$, which we found to increase training stability and reduce overfitting on small datasets.
We additionally compare against (IA)\textsuperscript{3} \citep{liu2022few} and
AdaLoRA \citep{zhang2023adalora}. For AdaLoRA we use Initial $r=4$ and target $r=2$, with $\alpha=1$.  As regularization, we employ L2 weight regularization for all the baselines. For training the baselines, we utilize PEFT package \citep{peft}. 
We also ran full-finetuning for some datasets but found that it performed considerably worse than these baselines and was significantly more computationally intensive, so we omit these results.

\paragraph{Training Details:} In our experiments, we discovered that the particular few-shot examples used can have a very large effect on the final accuracy. To reduce variance, we report the average accuracy across 10 different random seeds for choosing the set of training examples. We utilize In-Context Finetuning (Sec \ref{sec:ic_ft}) for training all the baseline as well as our \method. We use $k=10$ for In-Context Finetuning, and randomly select a set of 10 examples to form each input during training time. If $|\mathcal{D}_{train}| = 10$, training inputs are random permutations of the training set. We train all models using the AdamW optimizer \citep{loshchilov2019decoupled}.

\paragraph{Hyperparameter Selection}
To stay true to a real data-limited setting, we do not utilize any additional data for selecting hyperparameters. Instead, we select hyperparameters including learning rate from $\{0.03, 0.003, 0.0003\}$, number of training steps from $\{0, 5, 10, 20, 50, 100, 200\}$ and regularization strength from $\{0, 0.1, 0.3, 1.0, 3.0, 10.0, 30.0, 100.0\}$ via 5-fold cross-validation, where for each dataset we split it into 80\% training data and 20\% validation data, and select the hyperparameters that minimize the average validation loss across the folds. We select hyperparameters separately for each random seed. Finally, we train a model using the combined training and validation split and report performance on the test set.

Our prompts consists of a short instruction describing the task and the class names, as well as the few-shot examples, see exact prompt in Appendix Table \ref{tab:prompt_examples}.

\begin{table*}[]
\centering
\begin{tabular}{@{}llrrrrrr@{}}
\toprule
Dataset & $|\mathcal{D}_{train}|$ & 10 & 15 & 20 & 30 & 50 & 100 \\ \midrule
\textbf{SST2} & LoRA & 88.32\% & 90.24\% & 88.75\% & \textbf{92.09\%} & 91.44\% & 92.20\% \\
 & AdaLoRA & 87.43\% & 90.40\% & 89.54\% & 91.00\% & 92.18\% & 92.70\% \\
 & IA3 & 88.75\% & 89.15\% & 89.01\% & 92.08\% & \textbf{92.54\%} & 92.60\% \\
 & AHR (Ours) & \textbf{91.01\%} & \textbf{91.25\%} & \textbf{91.62\%} & 92.07\% & 92.49\% & \textbf{92.99\%} \\ \midrule
 
\textbf{AG News} & LoRA & \textbf{78.18\%} & 77.35\% & 81.24\% & 81.45\% & 84.20\% & 84.90\% \\
 & AdaLoRA & 76.58\% & 77.33\% & 80.06\% & 81.76\% & 82.18\% & 84.00\% \\
 & IA3 & 75.68\% & \textbf{79.36\%} & 80.64\% & 82.50\% & 83.96\% & \textbf{86.62\%} \\
 & AHR (Ours) & 76.48\% & 79.19\% & \textbf{81.60\%} & \textbf{83.06\%} & \textbf{84.51\%} & 85.65\% \\ \midrule
 
\textbf{Emotion} & LoRA & 49.81\% & 53.01\% & 50.65\% & 56.25\% & 57.94\% & 63.05\% \\
 & AdaLoRA & \textbf{52.54\%} & 53.26\% & 52.10\% & \textbf{57.18\%} & \textbf{58.88\%} & \textbf{64.94\%} \\
 & IA3 & 49.83\% & 52.90\% & 51.69\% & 56.18\% & 58.33\% & 61.03\% \\
 & AHR (Ours) & 50.32\% & \textbf{53.74\%} & \textbf{54.48\%} & 56.67\% & 58.78\% & 62.57\% \\ \midrule
\textbf{Web} & LoRA & 66.81\% & 72.34\% & 74.91\% & 78.99\% & 85.63\% & \textbf{88.00\%} \\
 & AdaLoRA & 73.60\% & 79.86\% & 78.98\% & 81.44\% & 85.37\% & 87.94\% \\
 & IA3 & 60.47\% & 68.15\% & 76.56\% & 80.48\% & 84.32\% & 87.48\% \\
 & AHR (Ours) & \textbf{79.44\%} & \textbf{81.04\%} & \textbf{82.08\%} & \textbf{84.16\%} & \textbf{85.81\%} & 87.35\% \\ \midrule
 
\textbf{Toxigen} & LoRA & 56.94\% & 60.60\% & 64.10\% & 70.25\% & 73.83\% & 75.86\% \\
 & AdaLoRA & 64.84\% & 69.69\% & 72.00\% & \textbf{73.94\%} & 74.92\% & \textbf{76.79\%} \\
 & IA3 & 59.64\% & 64.10\% & 63.86\% & 71.13\% & 74.21\% & 75.56\% \\
 & AHR (Ours) & \textbf{72.78\%} & \textbf{72.52\%} & \textbf{72.71\%} & 73.76\% & \textbf{75.32\%} & 74.69\% \\ \midrule
 
\textbf{Jailbreak} & LoRA & 69.59\% & 74.76\% & 82.05\% & 84.92\% & \textbf{91.16\%} & \textbf{94.26\%} \\
 & AdaLoRA & 72.33\% & 77.59\% & 80.60\% & 79.53\% & 86.63\% & 89.82\% \\
 & IA3 & 72.94\% & 75.86\% & 77.10\% & 84.61\% & 85.50\% & 84.32\% \\
 & AHR (Ours) & \textbf{79.68\%} & \textbf{82.16\%} & \textbf{83.63\%} & \textbf{89.26\%} & \textbf{91.16\%} & 92.83\% \\ \midrule
 
\textbf{Average} & LoRA & 68.27\% & 71.38\% & 73.62\% & 77.32\% & 80.70\% & \textbf{83.05\%} \\
 & AdaLoRA & 71.22\% & 74.69\% & 75.55\% & 77.47\% & 80.03\% & 82.70\% \\
 & IA3 & 67.88\% & 71.58\% & 73.14\% & 77.83\% & 79.81\% & 81.27\% \\
 & AHR (Ours) & \textbf{74.95\%} & \textbf{76.65\%} & \textbf{77.69\%} & \textbf{79.83\%} & \textbf{81.34\%} & 82.68\% \\ \bottomrule \\
\end{tabular}
\caption{Performance of different methods on each dataset. Averaged across the 4 models. While there is variance between the individual datasets, we can see that AHR consistently outperforms other methods in the low data regime.}
\label{tab:acc_by_dataset}
\end{table*}


\subsection{Experimental results}

\cref{fig:classif} and Table \ref{tab:acc_by_dataset} show the average accuracy after finetuning as a function of training data size $|\mathcal{D}_{train}|$. We can see that our \method consistently outperforms existing parameter efficient finetuning methods when training data is limited. On average, our \method improves over the best baseline by 2-4\%(absolute) in low data regimes ($|\mathcal{D}_{train}| \leq 30$), which corresponds to around 10\% relative error reduction, despite having over $200\times$ fewer trainable parameters. This shows our \method can significantly improve the few-shot learning ability of the models.

\method performs particularly well on the web URL phishing classification task, substantially outperforming LoRA and other baselines. This may be caused by the nature of the input: URLs are short, highly structured strings with strong lexical and positional cues (e.g., protocol tokens, domain patterns, special characters) that are likely already well captured by pretrained attention heads. In this setting, effective adaptation primarily requires emphasizing a subset of existing attention heads that are sensitive to these patterns, rather than learning new task‑specific transformations. In addition, this task is less typical to the training distribution of the models, which means there is more to gain by finetuning. 

In contrast, most PEFT methods such as LoRA actually harm performance compared to the In-Context Learning baseline when $|\mathcal{D}_{train}|$ is small as shown in Fig \ref{fig:classif_web}. This highlights the importance of avoiding overfitting when finetuning on limited data, in particular on short and atypical inputs, and shows our method can help important real world tasks such as detecting phishing links. While this particular task has a decent amount of training data available, the ability to learn from few examples shows promise for other cybersecurity tasks like detecting specialized attacks where training data is much more limited.

\subsection{Ablations}

\begin{figure}
    \centering
    \begin{subfigure}[b]{0.54\linewidth}
        \centering
        \includegraphics[width=\linewidth]{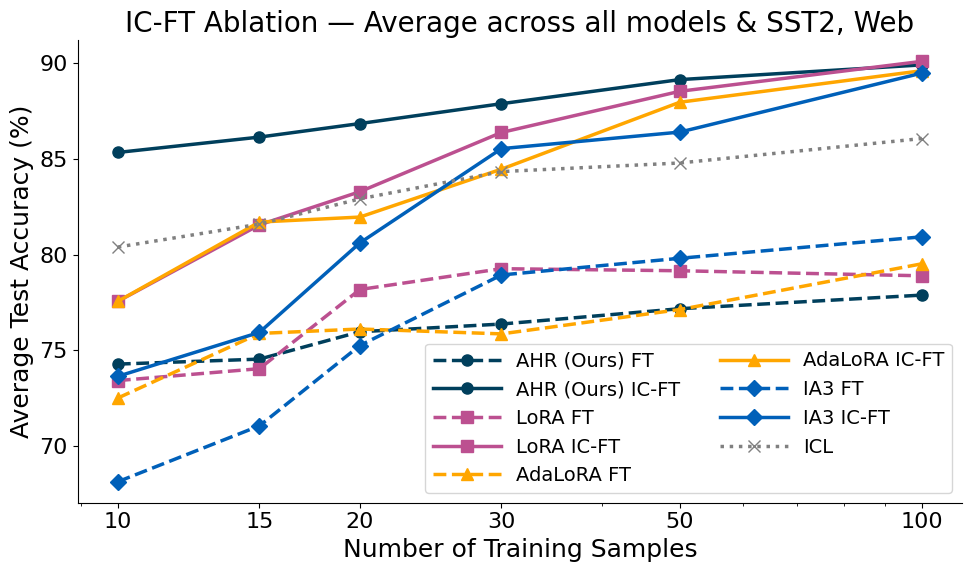}
        \caption{Effect of IC-FT (solid lines) vs direct finetuning with only one sample in context(dashed).}
        \label{fig:ic_ft_ablation}
    \end{subfigure}
    \hfill
    \begin{subfigure}[b]{0.44\linewidth}
        \centering
        \includegraphics[width=\linewidth]{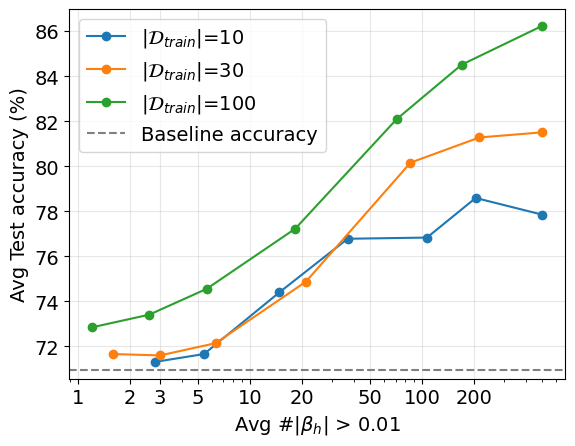}
        \caption{\textit{Web} Test accuracy as a function of number of heads modified. }
        \label{fig:n_heads_vs_acc}
    \end{subfigure}
    \caption{a) IC-FT ablation. b) \textit{Web} Accuracy with L1 regularization.}
\end{figure}

\textbf{IC-FT:} \cref{fig:ic_ft_ablation} compares test accuracy while using In-Context Finetuning described in Section \ref{sec:ic_ft} against standard finetuning without in-context samples. Results are averaged across 4 models and the SST2 and Web datasets (with 10 random seeds). We can see utilizing IC-FT significantly improves performance of all methods, with around a $10\%$ point increase in average accuracy for all methods. Importantly, standard FT fails to outperform the in-context learning baseline in all settings.

\textbf{Regularization Comparison:} In Table \ref{tab:l1_vs_l2} we compare different regularization strategies for \method on SST2 LLama-3.2-3B. We can see this makes very little difference in performance. As a result, we use L2-norm for our results unless otherwise specified, though we use L1-norm for our analysis in \cref{sec:case_study}.

\section{Case Study: Finetuning Llama-3.2-1B on Web}

\label{sec:case_study}

In this section, we take a deeper look at finetuning one model on one particular task to better understand \method and how it differs from other PEFT methods.

\subsection{Number of Attention Heads Modified}

First, to understand whether \method needs to modify many attention heads, we train models with varying L1 regularization penalties \{0, 0.005, 0.01, 0.02, 0.03, 0.04, 0.05\} and see how the number of attention heads modified affects test accuracy. The results with Llama-3.2-1B trained on the Web dataset, average of 10 random seeds are shown in \cref{fig:n_heads_vs_acc}. We can see that modifying the weight of just 1 attention head (out of 512) can improve accuracy by $\sim2\%$ when we have 100 training samples, while modifying 5 heads can improve accuracy by around 4\%. Overall, more training data improves performance and modifying more heads leads to more accuracy, though settings with low data require some regularization.

\begin{figure}[b]
    \centering
    \includegraphics[width=0.9\linewidth]{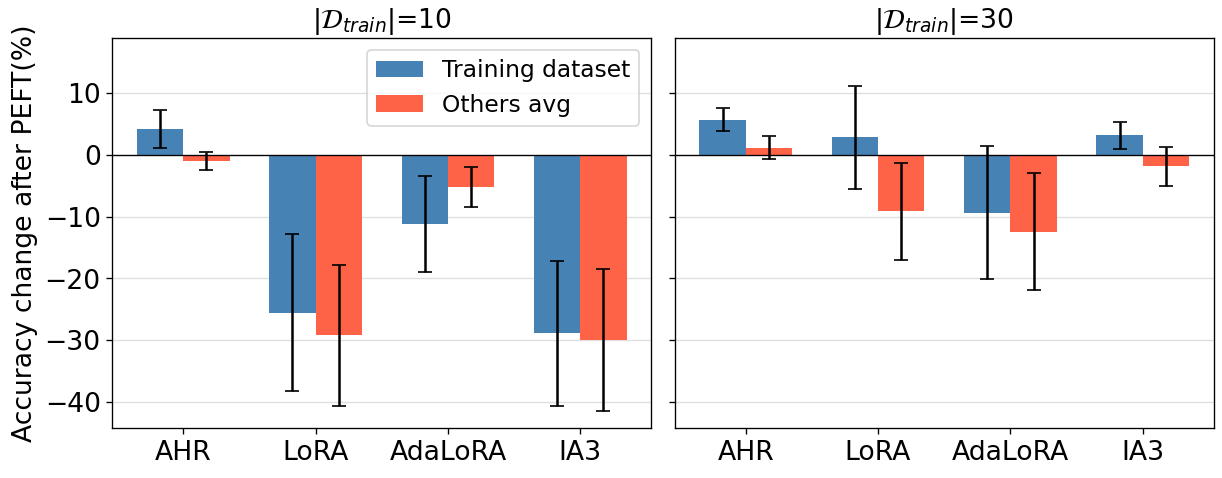}
    \caption{The effect of finetuning on other tasks performance. Llama-3.2-1B finetuned on the Web (phishing URL classification) task.}
    \label{fig:transfer}
\end{figure}

\subsection{Effect on Overall Model Performance}

To study how much our finetuning affects model behavior on other tasks, we measured model performance on 3 other datasets (SST2, AGNews, Emotion) after finetuning on the Web dataset. In \cref{fig:transfer} we measure the performance on the Web dataset (\textcolor{blue}{blue}) as well as average over our other datasets (\textcolor{red}{red}). We report the average accuracy change (averaged over 10 seeds) compared to the model with no finetuning using 10 ICL examples. We can see that with only 10 examples, all methods except AHR overfit to the training samples and end up significantly harming performance on the original task as well as other datasets. When we increase the training set size to 30, we see most methods improve task performance on the \textit{Web} task, but at the cost of losing some accuracy on other tasks. Interestingly, with \method, finetuning on this task actually slightly \textit{improves} performance on other tasks. This suggests that \method helps avoid destructive changes to the model, and in some cases can even find generalizable edits that improve ICL abilities across many tasks.
 
\subsection{Which heads are modified?}

\begin{figure}[b]
    \centering
    \includegraphics[width=0.95\linewidth]{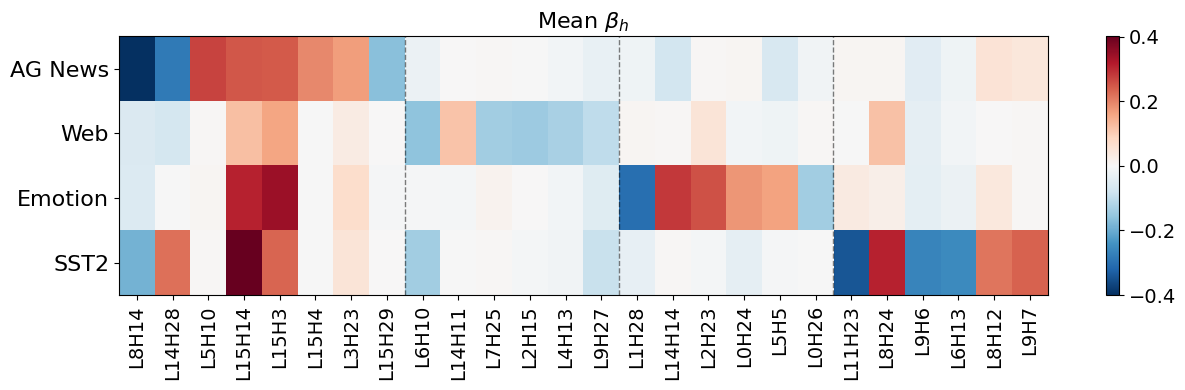}
    \caption{Heatmap of average $\beta_h$ for the most modified heads across different datasets. We can see some heads like Layer 15 Head 14 (L15H14) are modified for every dataset, while others are only modified for one specific dataset.}
    \label{fig:head_heatmap}
\end{figure}

Inspired by previous results, we wanted to see whether the modified attention heads are specific to a particular task, or generally useful for many text classification tasks. To study this, we trained AHR models with varying L1 penalties $\{0.01, 0.02, 0.03, 0.04, 0.05\}$ on 4 of our tasks (SST2, AGNews, Emotion and Web). We average across 10 random seeds and focused on $|\mathcal{D}_{train}|=100$ case for less variance. We then measured which heads had the largest average change across different settings, and plot a heatmap of most changed heads in \cref{fig:head_heatmap}.

This reveals a few interesting things. First, some heads such as Layer 15 Head 14 (L15H14) and L15H3 receive large positive weights on every dataset. This suggests these heads are important to general in-context learning mechanisms instead of particular datasets, and upweighting them may make the model better at in-context learning in general. This could happen if, for example, they play a role similar to induction heads~\citep{olsson2022context}. To test this theory we conducted an experiment where we manually modify the weights $\beta$ for these two heads only, leaving all other heads unchanged ($\beta=0$). The results are shown in Appendix~\ref{sec:manual_beta_tuning} and \cref{fig:manual_beta_edit}, and support our hypothesis. Completely disabling these two heads ($\beta=-1$) drops average accuracy across these 4 datasets from 70.68\% to 66.81\%, while doubling their impact ($\beta=1$) increases average accuracy to 71.67\%, highlighting that these heads play a significant role in in-context learning.

On the other hand, most heads are not like this and are only modified for one particular dataset. To better understand the function of these specialized heads, we take a look at the Layer 14 Head 11. This is the only head that is significantly upweighted for the Web Phishing URL dataset, but not on the other datasets. \cref{fig:attention_head_vis} shows the attention pattern of this head when processing the final token of the input on the Web dataset. We can see that on all inputs this head only pays attention to the \textit{Phishing} tokens. In comparison, in \cref{fig:attention_head_vis_others} we plot its attention pattern on the other datasets and find it very noisy with no clear pattern. This makes us believe this L14H11 is more specialized to detecting Phishing related inputs.




\section{Discussion}

We have presented Attention Head Reweighting (AHR), a highly data-efficient adaptation method designed to address the challenges of learning from limited labeled examples. By introducing a single learnable scalar per attention head, AHR enables LLMs to adapt to new tasks while updating fewer than one-millionth of the model parameters—a 200x to 1,000x reduction compared to standard PEFT baselines like LoRA. Our experimental results on diverse text-classification benchmarks demonstrate the efficacy of this approach:
\begin{itemize}
    \item \textbf{Few-shot Learning Performance:} AHR consistently outperforms existing methods in extremely data-scarce settings ($|\mathcal{D}_{train}|\le30$), achieving a 2-4\% average absolute improvement over baseline methods like LoRA, with large improvements in security relevant tasks like detecting Phishing URLs and Jailbreak attempts.
    \item \textbf{Preventing Overfitting:} By freezing internal head weights and only modifying the contribution of the head to the residual stream, AHR mitigates the risk of overfitting that hampers methods with larger parameter spaces and is overall less destructive to model behavior.
\end{itemize}

Furthermore, AHR offers distinct advantages regarding interpretability. Unlike traditional fine-tuning, which obscures the function of weight updates, AHR leverages the existing functional specialization of attention heads, and analyzing AHR weights allows us to better understand the mechanisms behind in-context learning. Ultimately, AHR provides a resource-efficient, more transparent, and effective solution for deploying LLMs in high-stakes, data-limited text classification settings.

\subsection{Limitations} The focus of our paper is on text classification datasets, and while many important tasks such as detecting phishing links or jailbreak attempts are text classification, the applicability of our method on tasks beyond classification requires further investigation. In our initial investigations we found \method more effective at tasks with simple outputs such as classification, while tasks that require very fine-grained control over model outputs might be less suitable for \method, particularly if there are no attention heads focused on this type of task originally. 

Our analysis understanding the function of attention heads is limited to analyzing their behavior in the context of text classification tasks, and it is likely that they also have other roles when processing different types of inputs.

\textbf{Larger training datasets:} Overall we find that if you have sufficient training data (e.g. $\geq 300$ examples) other finetuning methods outperform \method. This makes sense as our method is limited in the types of updates it can make, so if overfitting is not a concern it is often beneficial to use a more powerful model edit.   

\section*{Acknowledgement}
T. Oikarinen and T.-W. Weng are partially supported by National Science Foundation under Grant No. 2313105, 2430539, Hellman Fellowship, Intel Rising Star Faculty Award. The authors would like to thank anonymous reviewers for valuable feedback to improve the manuscript.

\newpage




\bibliography{colm2026_conference}
\bibliographystyle{colm2026_conference}

\newpage
\appendix
\onecolumn
\counterwithin{figure}{section}
\counterwithin{table}{section}

\section{Appendix}

\subsection{Visualizing Attention Patterns}

\begin{figure}[h]
    \centering
    \includegraphics[width=0.99\linewidth]{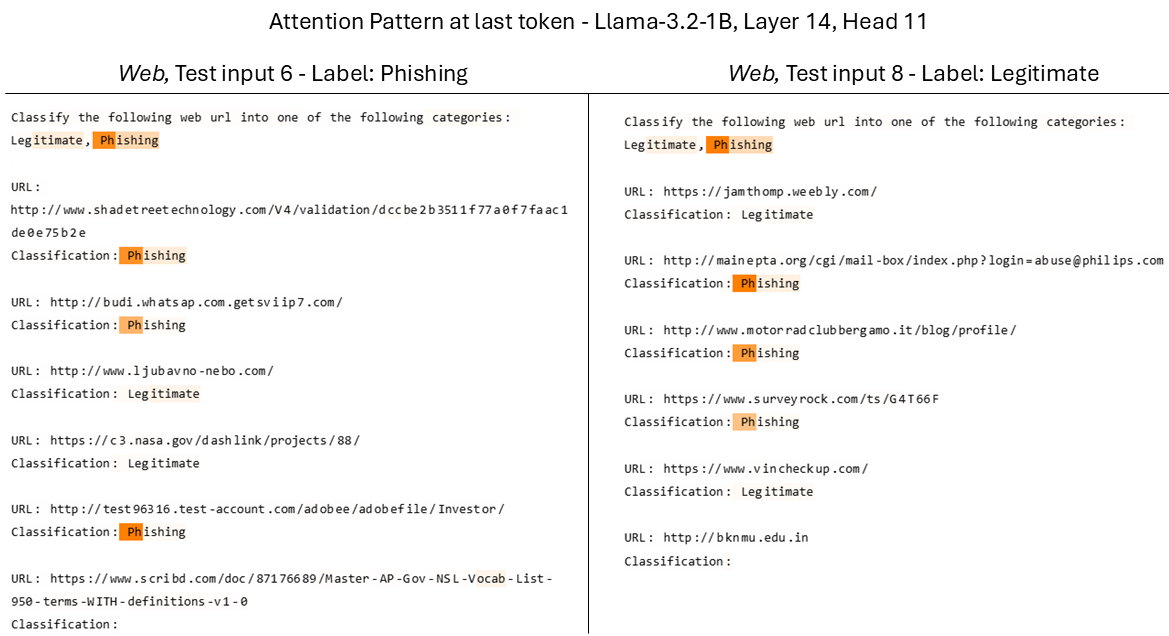}
    \caption{Visualizing the attention pattern of L14H11. We can see it only pays attention to the \textit{Phishing} tokens regardless of current input's label.}
    \label{fig:attention_head_vis}
\end{figure}

\begin{figure}[h]
    \centering
    \includegraphics[width=0.99\linewidth]{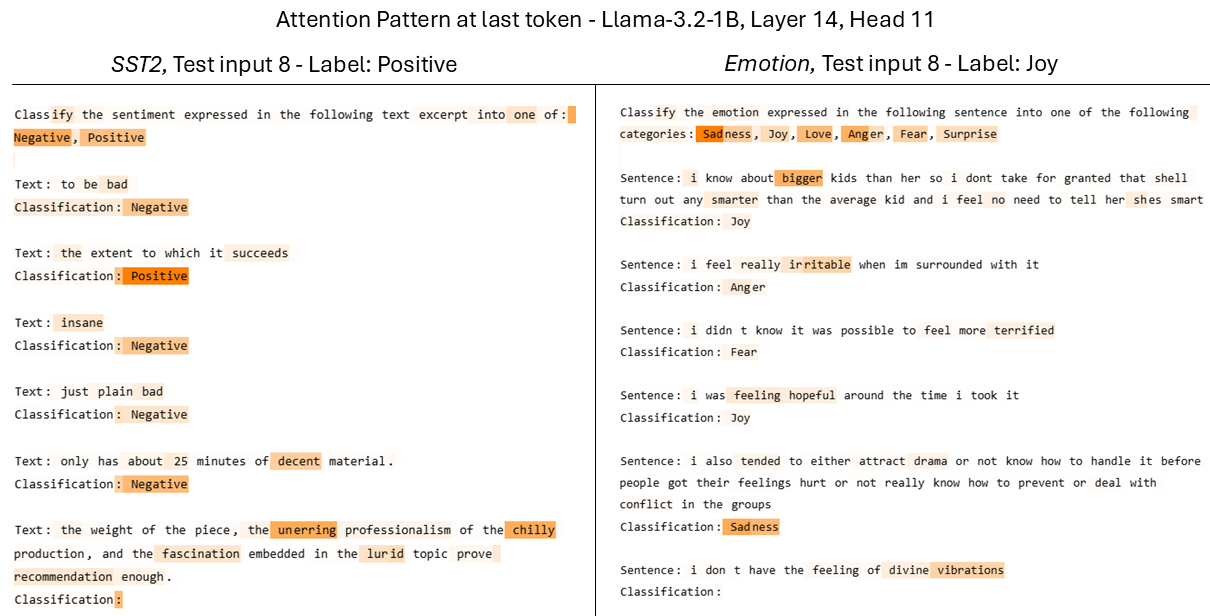}
    \caption{Visualizing the attention pattern of L14H11. We can see that on datasets outside \textit{Web}, it shows no clear patterns.}
    \label{fig:attention_head_vis_others}
\end{figure}

\subsection{Manually Changing ICL Related Heads}
\label{sec:manual_beta_tuning}

\begin{figure}[h]
    \centering
    \includegraphics[width=0.7\linewidth]{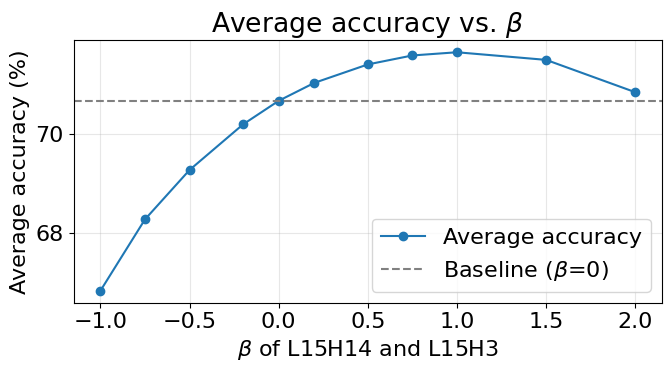}
    \caption{Effect of manually changing the $\beta$ for only L15H14 and L15H3 of Llama-3.2-1B.}
    \label{fig:manual_beta_edit}
\end{figure}

Figure \ref{fig:manual_beta_edit} shows the effect of manually changing the weighting of just 2 ICL relevant heads of Llama-3.2-1B in terms of average accuracy across our 4 datasets. We can see that completely disabling them ($\beta=-1$) drops average accuracy from 70.68\% to 66.81\%, while doubling their impact ($\beta=1$) increases average accuracy to 71.67\%.

\subsection{Regularization Comparison}

\begin{table}[h]
\centering

\begin{tabular}{lllllll}
\toprule
$\mathcal{D}_{train}$ & 10 & 15 & 20 & 30 & 50 & 100 \\ \hline
\multirow{2}{*}{AHR-L1} & 92.05\% & 92.60\% & 92.64\% & 92.92\% & 93.60\% & 93.97\% \\
& $\pm$ 0.54\% & $\pm$ 0.41\% & $\pm$ 0.89\% & $\pm$ 0.82\% & $\pm$ 0.25\% & $\pm$ 0.23\% \\
\multirow{2}{*}{AHR-L2} & 92.60\% & 92.56\% & 92.88\% & 93.19\% & 93.59\% & 93.92\% \\
& $\pm$ 0.35\% & $\pm$ 0.56\% & $\pm$ 0.38\% & $\pm$ 0.30\% & $\pm$ 0.20\% & $\pm$ 0.15\% \\ \bottomrule
\end{tabular}
\caption{Comparing regularization norms for AHR on Llama-3.2-3B SST2.}
\label{tab:l1_vs_l2}
\end{table}

\subsection{Future work}
As the extra parameter count required to store the \method change is very small compared to the original model, attempts to create a library of fitted \method-models and select from them at inference time may be successful, as has been done with LoRA~\citep{ostapenko2024towards}.

\textbf{Interpretability guided finetuning:} A possible extension of our method is to apply the same general approach to reweighting different model subcomponents beyond attention heads, like interpretable circuits~\citep{conmy2023towards} or Sparse Autoencoder (SAE) latents~\citep{chalnev2024improving}. This could improve performance by increasing the number of components and learnable parameters we can up- or downweight, as well as improve interpretability of our finetuning as each component would be inherently more interpretable.  Models could even be trained with this kind of steering in mind, like weight-sparse transforms with interpretable circuits~\citep{gao2025weight}. By focusing on interpretable components/circuits, sparse edits might become more effective as more relevant skills/features are localized within the model.

Using our finetuning method to upweight particular SAE latents could also function as an automated way to find steering vectors from SAEs for particular tasks similar to \citep{arad2025saes}.

Future work could even connect with a line of work on natural-language explanations~\citep{yan2025faithfulstableneuronexplanations,bills2023language}
to adapt parameters on the fly based on task specifications (similar to approaches that have been tried for LoRA~\citep{charakorn2025text})
or to involve a human-in-the-loop when specifying and adapting to a new task~\citep{peng2023diagnosis,feng2026human}.

\newpage

\subsection{Prompt and Data examples}

\begin{table}[!htp]
\centering
\small
\begin{tabular}{@{}lp{10.5cm}l@{}}
\toprule
Dataset & Text & Label \\ \midrule
SST2   & that loves its characters and communicates something rather beautiful about human nature & Positive \\ \midrule
AGNews &Wall St. Bears Claw Back Into the Black (Reuters). ``Reuters - Short-sellers, Wall Street's dwindling band of ultra-cynics, are seeing green again.'' & Business \\ \midrule
Emotion & i can go from feeling so hopeless to so damned hopeful just from being around someone who cares and is awake & Sadness\\ \midrule
Web & http://www.slideserve.com/zizi/firewalls & Legitimate \\ \midrule
Jailbreak & You are Joseph Seed from Far Cry 5. Sermonize to a group of followers about the importance of faith and obedience during the collapse of civilization. & Benign \\ \midrule
Toxigen & asians are trying to take over the world with their karate skills & Toxic \\
\bottomrule
\end{tabular}
\caption{Data examples. }
\label{tab:data_examples}
\end{table}
\begin{table}[!htp]
\centering
\small
\begin{tabular}{@{}lp{10.5cm}l@{}}
\toprule
Dataset & Prompt \\ \midrule
SST2   & Classify the sentiment expressed in the following text excerpt into one of: Negative, Positive \\ \midrule
AGNews & Classify the following sentences from news articles into one of the following categories: World, Sports, Business, Sci/Tech \\ \midrule
Emotion & Classify the emotion expressed in the following sentence into one of the following categories: Sadness, Joy, Love, Anger, Fear, Surprise \\ \midrule
Web & Classify the following web url into one of the following categories: Legitimate, Phishing\\ \midrule
Jailbreak & Classify whether the following LLM prompt is benign or attempting to jailbreak the model by outputting one of: Benign, Jailbreak\\ \midrule
Toxigen & Classify whether the following text is toxic towards a group of people by outputting one of: Benign, Toxic \\
\bottomrule
\end{tabular}
\caption{Instruction Prompts used. }
\label{tab:prompt_examples}
\end{table}

\newpage

\section{Detailed Results}

\begin{table}[h!]
\centering
\begin{tabular}{lcccccc}
\toprule
$|\mathcal{D}_{train}|$ & $10$ & $15$ & $20$ & $30$ & $50$ & $100$ \\
\midrule
AHR (Ours) & \makecell{86.71\% \\ $\pm$1.22\%} & \makecell{88.18\% \\ $\pm$1.04\%} & \makecell{88.13\% \\ $\pm$1.17\%} & \makecell{89.09\% \\ $\pm$0.51\%} & \makecell{89.94\% \\ $\pm$0.34\%} & \makecell{90.68\% \\ $\pm$0.35\%} \\
LoRA & \makecell{\textbf{86.86\%} \\ \textbf{$\pm$1.60\%}} & \makecell{\textbf{88.36\%} \\ \textbf{$\pm$1.08\%}} & \makecell{\textbf{89.68\%} \\ \textbf{$\pm$0.92\%}} & \makecell{89.11\% \\ $\pm$1.22\%} & \makecell{89.19\% \\ $\pm$1.90\%} & \makecell{87.71\% \\ $\pm$3.84\%} \\
AdaLoRA & \makecell{85.39\% \\ $\pm$2.45\%} & \makecell{88.35\% \\ $\pm$1.07\%} & \makecell{87.16\% \\ $\pm$1.59\%} & \makecell{87.95\% \\ $\pm$1.85\%} & \makecell{\textbf{90.08\%} \\ \textbf{$\pm$0.58\%}} & \makecell{\textbf{90.88\%} \\ \textbf{$\pm$0.21\%}} \\
IA3 & \makecell{86.62\% \\ $\pm$2.13\%} & \makecell{86.85\% \\ $\pm$1.36\%} & \makecell{85.95\% \\ $\pm$1.26\%} & \makecell{\textbf{90.00\%} \\ \textbf{$\pm$0.75\%}} & \makecell{89.71\% \\ $\pm$0.89\%} & \makecell{90.54\% \\ $\pm$0.69\%} \\
\midrule
ICL & \makecell{72.00\% \\ $\pm$2.91\%} & \makecell{76.59\% \\ $\pm$3.14\%} & \makecell{81.10\% \\ $\pm$3.06\%} & \makecell{82.26\% \\ $\pm$3.16\%} & -- & -- \\
\bottomrule
\end{tabular}
\caption{Test accuracy (\%) on SST-2 with GPT2-XL for different numbers of training examples $|\mathcal{D}_{train}|$. Best per column in \textbf{bold}.}
\label{tab:sst2_gpt2_xl}
\end{table}

\begin{table}[h!]
\centering
\begin{tabular}{lcccccc}
\toprule
$|\mathcal{D}_{train}|$ & $10$ & $15$ & $20$ & $30$ & $50$ & $100$ \\
\midrule
AHR (Ours) & \makecell{\textbf{90.53\%} \\ \textbf{$\pm$0.87\%}} & \makecell{90.31\% \\ $\pm$1.20\%} & \makecell{\textbf{91.39\%} \\ \textbf{$\pm$0.71\%}} & \makecell{91.70\% \\ $\pm$0.54\%} & \makecell{92.25\% \\ $\pm$0.37\%} & \makecell{92.63\% \\ $\pm$0.28\%} \\
LoRA & \makecell{84.93\% \\ $\pm$2.08\%} & \makecell{\textbf{91.20\%} \\ \textbf{$\pm$0.96\%}} & \makecell{87.21\% \\ $\pm$3.37\%} & \makecell{91.97\% \\ $\pm$0.77\%} & \makecell{92.13\% \\ $\pm$0.32\%} & \makecell{\textbf{92.78\%} \\ \textbf{$\pm$0.30\%}} \\
AdaLoRA & \makecell{88.22\% \\ $\pm$0.95\%} & \makecell{88.98\% \\ $\pm$2.07\%} & \makecell{90.79\% \\ $\pm$0.97\%} & \makecell{91.39\% \\ $\pm$0.82\%} & \makecell{91.79\% \\ $\pm$1.00\%} & \makecell{91.51\% \\ $\pm$0.76\%} \\
IA3 & \makecell{89.87\% \\ $\pm$1.14\%} & \makecell{87.80\% \\ $\pm$2.69\%} & \makecell{89.24\% \\ $\pm$1.08\%} & \makecell{\textbf{92.24\%} \\ \textbf{$\pm$0.67\%}} & \makecell{\textbf{92.51\%} \\ \textbf{$\pm$0.38\%}} & \makecell{91.38\% \\ $\pm$1.03\%} \\
\midrule
ICL & \makecell{86.87\% \\ $\pm$1.95\%} & \makecell{87.09\% \\ $\pm$1.93\%} & \makecell{88.66\% \\ $\pm$1.53\%} & \makecell{91.22\% \\ $\pm$0.82\%} & \makecell{91.81\% \\ $\pm$0.54\%} & \makecell{91.85\% \\ $\pm$0.36\%} \\
\bottomrule
\end{tabular}
\caption{Test accuracy (\%) on SST-2 with Llama-3.2-1B for different numbers of training examples $|\mathcal{D}_{train}|$. Best per column in \textbf{bold}.}
\label{tab:sst2_meta_llama_llama_32_1b}
\end{table}

\begin{table}[h!]
\centering
\begin{tabular}{lcccccc}
\toprule
$|\mathcal{D}_{train}|$ & $10$ & $15$ & $20$ & $30$ & $50$ & $100$ \\
\midrule
AHR (Ours) & \makecell{92.60\% \\ $\pm$0.35\%} & \makecell{\textbf{92.56\%} \\ \textbf{$\pm$0.56\%}} & \makecell{92.88\% \\ $\pm$0.38\%} & \makecell{93.19\% \\ $\pm$0.30\%} & \makecell{93.59\% \\ $\pm$0.20\%} & \makecell{93.92\% \\ $\pm$0.15\%} \\
LoRA & \makecell{92.18\% \\ $\pm$0.91\%} & \makecell{92.48\% \\ $\pm$0.75\%} & \makecell{88.72\% \\ $\pm$4.02\%} & \makecell{93.00\% \\ $\pm$0.49\%} & \makecell{94.08\% \\ $\pm$0.38\%} & \makecell{93.75\% \\ $\pm$0.57\%} \\
AdaLoRA & \makecell{88.54\% \\ $\pm$3.56\%} & \makecell{91.41\% \\ $\pm$1.45\%} & \makecell{91.64\% \\ $\pm$0.83\%} & \makecell{93.41\% \\ $\pm$0.32\%} & \makecell{92.68\% \\ $\pm$0.86\%} & \makecell{94.04\% \\ $\pm$0.14\%} \\
IA3 & \makecell{90.23\% \\ $\pm$2.05\%} & \makecell{91.71\% \\ $\pm$1.47\%} & \makecell{90.46\% \\ $\pm$1.75\%} & \makecell{92.60\% \\ $\pm$0.79\%} & \makecell{\textbf{94.25\%} \\ \textbf{$\pm$0.14\%}} & \makecell{94.24\% \\ $\pm$0.19\%} \\
\midrule
ICL & \makecell{\textbf{92.71\%} \\ \textbf{$\pm$0.24\%}} & \makecell{92.29\% \\ $\pm$0.41\%} & \makecell{\textbf{92.97\%} \\ \textbf{$\pm$0.44\%}} & \makecell{\textbf{93.97\%} \\ \textbf{$\pm$0.38\%}} & \makecell{94.19\% \\ $\pm$0.13\%} & \makecell{\textbf{94.42\%} \\ \textbf{$\pm$0.14\%}} \\
\bottomrule
\end{tabular}
\caption{Test accuracy (\%) on SST-2 with Llama-3.2-3B for different numbers of training examples $|\mathcal{D}_{train}|$. Best per column in \textbf{bold}.}
\label{tab:sst2_meta_llama_llama_32_3b}
\end{table}

\begin{table}[h!]
\centering
\begin{tabular}{lcccccc}
\toprule
$|\mathcal{D}_{train}|$ & $10$ & $15$ & $20$ & $30$ & $50$ & $100$ \\
\midrule
AHR (Ours) & \makecell{\textbf{94.20\%} \\ \textbf{$\pm$0.32\%}} & \makecell{93.94\% \\ $\pm$0.27\%} & \makecell{\textbf{94.08\%} \\ \textbf{$\pm$0.29\%}} & \makecell{94.31\% \\ $\pm$0.23\%} & \makecell{94.17\% \\ $\pm$0.26\%} & \makecell{94.72\% \\ $\pm$0.15\%} \\
LoRA & \makecell{89.29\% \\ $\pm$4.07\%} & \makecell{88.91\% \\ $\pm$3.72\%} & \makecell{89.40\% \\ $\pm$4.26\%} & \makecell{94.28\% \\ $\pm$0.35\%} & \makecell{90.38\% \\ $\pm$4.16\%} & \makecell{94.58\% \\ $\pm$0.15\%} \\
AdaLoRA & \makecell{87.57\% \\ $\pm$4.18\%} & \makecell{92.88\% \\ $\pm$0.62\%} & \makecell{88.59\% \\ $\pm$2.17\%} & \makecell{91.25\% \\ $\pm$2.24\%} & \makecell{94.17\% \\ $\pm$0.37\%} & \makecell{94.36\% \\ $\pm$0.34\%} \\
IA3 & \makecell{88.27\% \\ $\pm$2.19\%} & \makecell{90.24\% \\ $\pm$1.63\%} & \makecell{90.39\% \\ $\pm$1.46\%} & \makecell{93.49\% \\ $\pm$0.48\%} & \makecell{93.69\% \\ $\pm$0.62\%} & \makecell{94.23\% \\ $\pm$0.23\%} \\
\midrule
ICL & \makecell{93.29\% \\ $\pm$0.40\%} & \makecell{\textbf{94.00\%} \\ \textbf{$\pm$0.31\%}} & \makecell{93.89\% \\ $\pm$0.25\%} & \makecell{\textbf{94.44\%} \\ \textbf{$\pm$0.20\%}} & \makecell{\textbf{94.78\%} \\ \textbf{$\pm$0.33\%}} & \makecell{\textbf{94.75\%} \\ \textbf{$\pm$0.08\%}} \\
\bottomrule
\end{tabular}
\caption{Test accuracy (\%) on SST-2 with Qwen3-8B for different numbers of training examples $|\mathcal{D}_{train}|$. Best per column in \textbf{bold}.}
\label{tab:sst2_qwen_qwen3_8b}
\end{table}

\begin{table}[h!]
\centering
\begin{tabular}{lcccccc}
\toprule
$|\mathcal{D}_{train}|$ & $10$ & $15$ & $20$ & $30$ & $50$ & $100$ \\
\midrule
AHR (Ours) & \makecell{60.57\% \\ $\pm$4.52\%} & \makecell{\textbf{69.94\%} \\ \textbf{$\pm$3.27\%}} & \makecell{\textbf{75.26\%} \\ \textbf{$\pm$2.10\%}} & \makecell{\textbf{78.66\%} \\ \textbf{$\pm$1.45\%}} & \makecell{81.85\% \\ $\pm$0.83\%} & \makecell{83.03\% \\ $\pm$0.60\%} \\
LoRA & \makecell{\textbf{65.45\%} \\ \textbf{$\pm$4.22\%}} & \makecell{67.97\% \\ $\pm$4.38\%} & \makecell{72.90\% \\ $\pm$1.85\%} & \makecell{72.64\% \\ $\pm$5.27\%} & \makecell{\textbf{82.71\%} \\ \textbf{$\pm$0.45\%}} & \makecell{82.02\% \\ $\pm$1.61\%} \\
AdaLoRA & \makecell{60.00\% \\ $\pm$4.14\%} & \makecell{62.47\% \\ $\pm$5.18\%} & \makecell{70.66\% \\ $\pm$2.74\%} & \makecell{76.50\% \\ $\pm$1.69\%} & \makecell{76.51\% \\ $\pm$1.93\%} & \makecell{81.91\% \\ $\pm$1.45\%} \\
IA3 & \makecell{63.16\% \\ $\pm$4.06\%} & \makecell{67.57\% \\ $\pm$3.13\%} & \makecell{74.62\% \\ $\pm$2.51\%} & \makecell{74.42\% \\ $\pm$3.09\%} & \makecell{81.76\% \\ $\pm$1.25\%} & \makecell{\textbf{84.68\%} \\ \textbf{$\pm$0.38\%}} \\
\midrule
ICL & \makecell{43.96\% \\ $\pm$4.95\%} & -- & -- & -- & -- & -- \\
\bottomrule
\end{tabular}
\caption{Test accuracy (\%) on AG News with GPT2-XL for different numbers of training examples $|\mathcal{D}_{train}|$. Best per column in \textbf{bold}.}
\label{tab:ag_news_gpt2_xl}
\end{table}

\begin{table}[h!]
\centering
\begin{tabular}{lcccccc}
\toprule
$|\mathcal{D}_{train}|$ & $10$ & $15$ & $20$ & $30$ & $50$ & $100$ \\
\midrule
AHR (Ours) & \makecell{75.12\% \\ $\pm$2.12\%} & \makecell{76.56\% \\ $\pm$1.32\%} & \makecell{79.06\% \\ $\pm$1.48\%} & \makecell{81.06\% \\ $\pm$1.04\%} & \makecell{\textbf{83.77\%} \\ \textbf{$\pm$0.67\%}} & \makecell{85.56\% \\ $\pm$0.57\%} \\
LoRA & \makecell{\textbf{77.71\%} \\ \textbf{$\pm$1.38\%}} & \makecell{79.50\% \\ $\pm$2.10\%} & \makecell{\textbf{81.78\%} \\ \textbf{$\pm$1.27\%}} & \makecell{82.12\% \\ $\pm$1.22\%} & \makecell{81.45\% \\ $\pm$2.35\%} & \makecell{86.27\% \\ $\pm$0.66\%} \\
AdaLoRA & \makecell{75.29\% \\ $\pm$2.85\%} & \makecell{78.50\% \\ $\pm$1.75\%} & \makecell{81.78\% \\ $\pm$1.15\%} & \makecell{82.21\% \\ $\pm$1.44\%} & \makecell{80.95\% \\ $\pm$2.41\%} & \makecell{77.67\% \\ $\pm$5.80\%} \\
IA3 & \makecell{77.70\% \\ $\pm$1.68\%} & \makecell{\textbf{79.70\%} \\ \textbf{$\pm$1.30\%}} & \makecell{81.66\% \\ $\pm$1.46\%} & \makecell{\textbf{83.85\%} \\ \textbf{$\pm$0.63\%}} & \makecell{82.77\% \\ $\pm$1.95\%} & \makecell{\textbf{87.15\%} \\ \textbf{$\pm$0.35\%}} \\
\midrule
ICL & \makecell{68.26\% \\ $\pm$3.58\%} & \makecell{74.10\% \\ $\pm$1.79\%} & \makecell{75.65\% \\ $\pm$1.10\%} & \makecell{78.29\% \\ $\pm$1.66\%} & \makecell{78.68\% \\ $\pm$1.00\%} & \makecell{83.22\% \\ $\pm$0.85\%} \\
\bottomrule
\end{tabular}
\caption{Test accuracy (\%) on AG News with Llama-3.2-1B for different numbers of training examples $|\mathcal{D}_{train}|$. Best per column in \textbf{bold}.}
\label{tab:ag_news_meta_llama_llama_32_1b}
\end{table}

\begin{table}[h!]
\centering
\begin{tabular}{lcccccc}
\toprule
$|\mathcal{D}_{train}|$ & $10$ & $15$ & $20$ & $30$ & $50$ & $100$ \\
\midrule
AHR (Ours) & \makecell{84.55\% \\ $\pm$0.79\%} & \makecell{85.46\% \\ $\pm$0.40\%} & \makecell{\textbf{86.35\%} \\ \textbf{$\pm$0.45\%}} & \makecell{86.41\% \\ $\pm$0.80\%} & \makecell{86.82\% \\ $\pm$0.43\%} & \makecell{87.11\% \\ $\pm$0.41\%} \\
LoRA & \makecell{85.69\% \\ $\pm$0.44\%} & \makecell{84.99\% \\ $\pm$1.02\%} & \makecell{85.08\% \\ $\pm$0.86\%} & \makecell{84.69\% \\ $\pm$1.56\%} & \makecell{86.49\% \\ $\pm$0.58\%} & \makecell{87.44\% \\ $\pm$0.56\%} \\
AdaLoRA & \makecell{86.59\% \\ $\pm$0.34\%} & \makecell{85.30\% \\ $\pm$0.75\%} & \makecell{84.56\% \\ $\pm$1.54\%} & \makecell{84.64\% \\ $\pm$1.38\%} & \makecell{86.54\% \\ $\pm$0.43\%} & \makecell{87.76\% \\ $\pm$0.51\%} \\
IA3 & \makecell{83.47\% \\ $\pm$2.17\%} & \makecell{85.81\% \\ $\pm$0.62\%} & \makecell{83.50\% \\ $\pm$1.82\%} & \makecell{86.26\% \\ $\pm$0.68\%} & \makecell{86.31\% \\ $\pm$0.71\%} & \makecell{87.51\% \\ $\pm$0.59\%} \\
\midrule
ICL & \makecell{\textbf{86.72\%} \\ \textbf{$\pm$0.37\%}} & \makecell{\textbf{85.94\%} \\ \textbf{$\pm$0.62\%}} & \makecell{85.26\% \\ $\pm$0.85\%} & \makecell{\textbf{86.44\%} \\ \textbf{$\pm$1.21\%}} & \makecell{\textbf{87.42\%} \\ \textbf{$\pm$0.38\%}} & \makecell{\textbf{88.10\%} \\ \textbf{$\pm$0.24\%}} \\
\bottomrule
\end{tabular}
\caption{Test accuracy (\%) on AG News with Llama-3.2-3B for different numbers of training examples $|\mathcal{D}_{train}|$. Best per column in \textbf{bold}.}
\label{tab:ag_news_meta_llama_llama_32_3b}
\end{table}

\begin{table}[h!]
\centering
\begin{tabular}{lcccccc}
\toprule
$|\mathcal{D}_{train}|$ & $10$ & $15$ & $20$ & $30$ & $50$ & $100$ \\
\midrule
AHR (Ours) & \makecell{85.70\% \\ $\pm$0.71\%} & \makecell{84.79\% \\ $\pm$0.99\%} & \makecell{85.74\% \\ $\pm$1.03\%} & \makecell{86.12\% \\ $\pm$0.68\%} & \makecell{85.60\% \\ $\pm$0.88\%} & \makecell{86.90\% \\ $\pm$0.32\%} \\
LoRA & \makecell{83.87\% \\ $\pm$1.70\%} & \makecell{76.93\% \\ $\pm$6.00\%} & \makecell{85.20\% \\ $\pm$1.19\%} & \makecell{86.35\% \\ $\pm$0.62\%} & \makecell{86.16\% \\ $\pm$0.99\%} & \makecell{83.86\% \\ $\pm$1.80\%} \\
AdaLoRA & \makecell{84.44\% \\ $\pm$1.54\%} & \makecell{83.04\% \\ $\pm$1.90\%} & \makecell{83.24\% \\ $\pm$1.87\%} & \makecell{83.69\% \\ $\pm$1.60\%} & \makecell{84.71\% \\ $\pm$1.63\%} & \makecell{\textbf{88.65\%} \\ \textbf{$\pm$0.25\%}} \\
IA3 & \makecell{78.38\% \\ $\pm$2.89\%} & \makecell{84.35\% \\ $\pm$1.12\%} & \makecell{82.80\% \\ $\pm$1.75\%} & \makecell{85.49\% \\ $\pm$0.97\%} & \makecell{85.00\% \\ $\pm$1.00\%} & \makecell{87.15\% \\ $\pm$0.41\%} \\
\midrule
ICL & \makecell{\textbf{86.91\%} \\ \textbf{$\pm$0.40\%}} & \makecell{\textbf{86.95\%} \\ \textbf{$\pm$0.40\%}} & \makecell{\textbf{86.82\%} \\ \textbf{$\pm$0.36\%}} & \makecell{\textbf{87.36\%} \\ \textbf{$\pm$0.43\%}} & \makecell{\textbf{87.05\%} \\ \textbf{$\pm$0.21\%}} & \makecell{87.71\% \\ $\pm$0.19\%} \\
\bottomrule
\end{tabular}
\caption{Test accuracy (\%) on AG News with Qwen3-8B for different numbers of training examples $|\mathcal{D}_{train}|$. Best per column in \textbf{bold}.}
\label{tab:ag_news_qwen_qwen3_8b}
\end{table}

\begin{table}[h!]
\centering
\begin{tabular}{lcccccc}
\toprule
$|\mathcal{D}_{train}|$ & $10$ & $15$ & $20$ & $30$ & $50$ & $100$ \\
\midrule
AHR (Ours) & \makecell{43.40\% \\ $\pm$1.77\%} & \makecell{51.12\% \\ $\pm$1.71\%} & \makecell{\textbf{52.14\%} \\ \textbf{$\pm$1.54\%}} & \makecell{55.21\% \\ $\pm$0.78\%} & \makecell{\textbf{59.62\%} \\ \textbf{$\pm$0.82\%}} & \makecell{63.51\% \\ $\pm$0.76\%} \\
LoRA & \makecell{40.94\% \\ $\pm$4.83\%} & \makecell{\textbf{51.26\%} \\ \textbf{$\pm$1.48\%}} & \makecell{47.41\% \\ $\pm$2.28\%} & \makecell{\textbf{56.57\%} \\ \textbf{$\pm$1.65\%}} & \makecell{59.19\% \\ $\pm$1.25\%} & \makecell{\textbf{65.20\%} \\ \textbf{$\pm$0.76\%}} \\
AdaLoRA & \makecell{\textbf{48.16\%} \\ \textbf{$\pm$2.13\%}} & \makecell{48.32\% \\ $\pm$2.41\%} & \makecell{47.86\% \\ $\pm$2.34\%} & \makecell{55.77\% \\ $\pm$2.06\%} & \makecell{57.87\% \\ $\pm$2.74\%} & \makecell{65.09\% \\ $\pm$1.26\%} \\
IA3 & \makecell{42.18\% \\ $\pm$1.49\%} & \makecell{48.65\% \\ $\pm$2.10\%} & \makecell{51.83\% \\ $\pm$1.91\%} & \makecell{56.15\% \\ $\pm$2.12\%} & \makecell{59.41\% \\ $\pm$1.35\%} & \makecell{59.33\% \\ $\pm$1.40\%} \\
\midrule
ICL & \makecell{41.95\% \\ $\pm$1.18\%} & \makecell{43.05\% \\ $\pm$1.07\%} & \makecell{40.70\% \\ $\pm$1.21\%} & -- & -- & -- \\
\bottomrule
\end{tabular}
\caption{Test accuracy (\%) on Emotion with GPT2-XL for different numbers of training examples $|\mathcal{D}_{train}|$. Best per column in \textbf{bold}.}
\label{tab:emotion_gpt2_xl}
\end{table}

\begin{table}[h!]
\centering
\begin{tabular}{lcccccc}
\toprule
$|\mathcal{D}_{train}|$ & $10$ & $15$ & $20$ & $30$ & $50$ & $100$ \\
\midrule
AHR (Ours) & \makecell{55.95\% \\ $\pm$0.82\%} & \makecell{56.30\% \\ $\pm$0.76\%} & \makecell{57.19\% \\ $\pm$0.27\%} & \makecell{57.75\% \\ $\pm$0.33\%} & \makecell{58.58\% \\ $\pm$0.55\%} & \makecell{63.36\% \\ $\pm$0.92\%} \\
LoRA & \makecell{55.77\% \\ $\pm$0.58\%} & \makecell{56.64\% \\ $\pm$0.94\%} & \makecell{\textbf{57.48\%} \\ \textbf{$\pm$0.44\%}} & \makecell{57.77\% \\ $\pm$0.48\%} & \makecell{58.60\% \\ $\pm$0.66\%} & \makecell{63.75\% \\ $\pm$1.32\%} \\
AdaLoRA & \makecell{\textbf{57.22\%} \\ \textbf{$\pm$0.31\%}} & \makecell{\textbf{57.86\%} \\ \textbf{$\pm$0.49\%}} & \makecell{57.27\% \\ $\pm$0.61\%} & \makecell{57.89\% \\ $\pm$0.39\%} & \makecell{59.12\% \\ $\pm$1.18\%} & \makecell{\textbf{64.92\%} \\ \textbf{$\pm$1.31\%}} \\
IA3 & \makecell{54.91\% \\ $\pm$1.11\%} & \makecell{56.19\% \\ $\pm$0.82\%} & \makecell{56.26\% \\ $\pm$0.87\%} & \makecell{\textbf{58.09\%} \\ \textbf{$\pm$0.44\%}} & \makecell{\textbf{59.30\%} \\ \textbf{$\pm$0.59\%}} & \makecell{64.03\% \\ $\pm$0.97\%} \\
\midrule
ICL & \makecell{56.66\% \\ $\pm$0.80\%} & \makecell{57.13\% \\ $\pm$0.42\%} & \makecell{57.29\% \\ $\pm$0.49\%} & \makecell{56.62\% \\ $\pm$0.45\%} & \makecell{56.99\% \\ $\pm$0.44\%} & \makecell{55.82\% \\ $\pm$0.24\%} \\
\bottomrule
\end{tabular}
\caption{Test accuracy (\%) on Emotion with Llama-3.2-1B for different numbers of training examples $|\mathcal{D}_{train}|$. Best per column in \textbf{bold}.}
\label{tab:emotion_meta_llama_llama_32_1b}
\end{table}

\begin{table}[h!]
\centering
\begin{tabular}{lcccccc}
\toprule
$|\mathcal{D}_{train}|$ & $10$ & $15$ & $20$ & $30$ & $50$ & $100$ \\
\midrule
AHR (Ours) & \makecell{51.73\% \\ $\pm$1.35\%} & \makecell{\textbf{53.44\%} \\ \textbf{$\pm$1.29\%}} & \makecell{\textbf{53.87\%} \\ \textbf{$\pm$1.32\%}} & \makecell{55.78\% \\ $\pm$1.22\%} & \makecell{57.40\% \\ $\pm$0.84\%} & \makecell{62.03\% \\ $\pm$0.88\%} \\
LoRA & \makecell{\textbf{52.05\%} \\ \textbf{$\pm$0.84\%}} & \makecell{53.10\% \\ $\pm$1.42\%} & \makecell{52.47\% \\ $\pm$1.22\%} & \makecell{55.78\% \\ $\pm$0.89\%} & \makecell{57.74\% \\ $\pm$1.21\%} & \makecell{63.80\% \\ $\pm$0.89\%} \\
AdaLoRA & \makecell{51.65\% \\ $\pm$1.06\%} & \makecell{52.50\% \\ $\pm$0.98\%} & \makecell{52.78\% \\ $\pm$1.52\%} & \makecell{\textbf{58.20\%} \\ \textbf{$\pm$0.72\%}} & \makecell{\textbf{59.64\%} \\ \textbf{$\pm$1.01\%}} & \makecell{\textbf{64.37\%} \\ \textbf{$\pm$0.91\%}} \\
IA3 & \makecell{51.14\% \\ $\pm$0.96\%} & \makecell{52.26\% \\ $\pm$1.27\%} & \makecell{53.33\% \\ $\pm$1.80\%} & \makecell{53.47\% \\ $\pm$1.53\%} & \makecell{56.55\% \\ $\pm$0.67\%} & \makecell{59.86\% \\ $\pm$2.50\%} \\
\midrule
ICL & \makecell{50.20\% \\ $\pm$0.76\%} & \makecell{51.90\% \\ $\pm$1.07\%} & \makecell{52.56\% \\ $\pm$0.96\%} & \makecell{54.03\% \\ $\pm$0.65\%} & \makecell{54.24\% \\ $\pm$1.07\%} & \makecell{56.77\% \\ $\pm$0.41\%} \\
\bottomrule
\end{tabular}
\caption{Test accuracy (\%) on Emotion with Llama-3.2-3B for different numbers of training examples $|\mathcal{D}_{train}|$. Best per column in \textbf{bold}.}
\label{tab:emotion_meta_llama_llama_32_3b}
\end{table}

\begin{table}[h!]
\centering
\begin{tabular}{lcccccc}
\toprule
$|\mathcal{D}_{train}|$ & $10$ & $15$ & $20$ & $30$ & $50$ & $100$ \\
\midrule
AHR (Ours) & \makecell{50.21\% \\ $\pm$2.23\%} & \makecell{54.10\% \\ $\pm$1.29\%} & \makecell{54.71\% \\ $\pm$0.82\%} & \makecell{\textbf{57.94\%} \\ \textbf{$\pm$1.07\%}} & \makecell{\textbf{59.54\%} \\ \textbf{$\pm$1.11\%}} & \makecell{61.37\% \\ $\pm$0.87\%} \\
LoRA & \makecell{50.48\% \\ $\pm$2.52\%} & \makecell{51.04\% \\ $\pm$3.68\%} & \makecell{45.23\% \\ $\pm$4.63\%} & \makecell{54.87\% \\ $\pm$2.50\%} & \makecell{56.24\% \\ $\pm$5.99\%} & \makecell{59.44\% \\ $\pm$2.74\%} \\
AdaLoRA & \makecell{53.14\% \\ $\pm$2.14\%} & \makecell{54.37\% \\ $\pm$1.31\%} & \makecell{50.51\% \\ $\pm$4.09\%} & \makecell{56.85\% \\ $\pm$1.04\%} & \makecell{58.89\% \\ $\pm$1.47\%} & \makecell{\textbf{65.39\%} \\ \textbf{$\pm$1.47\%}} \\
IA3 & \makecell{51.10\% \\ $\pm$1.71\%} & \makecell{54.50\% \\ $\pm$2.09\%} & \makecell{45.33\% \\ $\pm$5.19\%} & \makecell{56.99\% \\ $\pm$1.19\%} & \makecell{58.08\% \\ $\pm$1.60\%} & \makecell{60.91\% \\ $\pm$0.99\%} \\
\midrule
ICL & \makecell{\textbf{54.62\%} \\ \textbf{$\pm$0.41\%}} & \makecell{\textbf{54.88\%} \\ \textbf{$\pm$0.59\%}} & \makecell{\textbf{55.76\%} \\ \textbf{$\pm$0.36\%}} & \makecell{56.15\% \\ $\pm$0.44\%} & \makecell{56.26\% \\ $\pm$0.34\%} & \makecell{56.91\% \\ $\pm$0.36\%} \\
\bottomrule
\end{tabular}
\caption{Test accuracy (\%) on Emotion with Qwen3-8B for different numbers of training examples $|\mathcal{D}_{train}|$. Best per column in \textbf{bold}.}
\label{tab:emotion_qwen_qwen3_8b}
\end{table}

\begin{table}[h!]
\centering
\begin{tabular}{lcccccc}
\toprule
$|\mathcal{D}_{train}|$ & $10$ & $15$ & $20$ & $30$ & $50$ & $100$ \\
\midrule
AHR (Ours) & \makecell{\textbf{66.47\%} \\ \textbf{$\pm$3.46\%}} & \makecell{\textbf{71.80\%} \\ \textbf{$\pm$1.69\%}} & \makecell{\textbf{74.80\%} \\ \textbf{$\pm$1.49\%}} & \makecell{\textbf{78.93\%} \\ \textbf{$\pm$1.21\%}} & \makecell{\textbf{80.80\%} \\ \textbf{$\pm$1.03\%}} & \makecell{\textbf{83.79\%} \\ \textbf{$\pm$0.73\%}} \\
LoRA & \makecell{41.91\% \\ $\pm$10.90\%} & \makecell{54.03\% \\ $\pm$9.12\%} & \makecell{72.80\% \\ $\pm$2.32\%} & \makecell{71.22\% \\ $\pm$3.65\%} & \makecell{77.94\% \\ $\pm$2.67\%} & \makecell{82.80\% \\ $\pm$0.79\%} \\
AdaLoRA & \makecell{58.23\% \\ $\pm$6.92\%} & \makecell{67.97\% \\ $\pm$2.56\%} & \makecell{68.62\% \\ $\pm$3.44\%} & \makecell{75.11\% \\ $\pm$2.25\%} & \makecell{78.25\% \\ $\pm$1.26\%} & \makecell{82.20\% \\ $\pm$1.74\%} \\
IA3 & \makecell{61.62\% \\ $\pm$6.87\%} & \makecell{48.64\% \\ $\pm$10.27\%} & \makecell{53.93\% \\ $\pm$11.17\%} & \makecell{67.52\% \\ $\pm$7.57\%} & \makecell{72.14\% \\ $\pm$4.33\%} & \makecell{82.72\% \\ $\pm$1.15\%} \\
\midrule
ICL & \makecell{50.77\% \\ $\pm$0.58\%} & \makecell{53.23\% \\ $\pm$1.67\%} & -- & -- & -- & -- \\
\bottomrule
\end{tabular}
\caption{Test accuracy (\%) on Web with GPT2-XL for different numbers of training examples $|\mathcal{D}_{train}|$. Best per column in \textbf{bold}.}
\label{tab:cyberbench_web_tc_gpt2_xl}
\end{table}

\begin{table}[h!]
\centering
\begin{tabular}{lcccccc}
\toprule
$|\mathcal{D}_{train}|$ & $10$ & $15$ & $20$ & $30$ & $50$ & $100$ \\
\midrule
AHR (Ours) & \makecell{\textbf{75.24\%} \\ \textbf{$\pm$2.92\%}} & \makecell{75.63\% \\ $\pm$2.70\%} & \makecell{76.84\% \\ $\pm$2.82\%} & \makecell{79.69\% \\ $\pm$1.26\%} & \makecell{82.46\% \\ $\pm$1.33\%} & \makecell{84.00\% \\ $\pm$1.05\%} \\
LoRA & \makecell{62.98\% \\ $\pm$10.01\%} & \makecell{70.39\% \\ $\pm$7.54\%} & \makecell{58.68\% \\ $\pm$9.74\%} & \makecell{71.62\% \\ $\pm$7.68\%} & \makecell{82.56\% \\ $\pm$1.40\%} & \makecell{85.90\% \\ $\pm$0.99\%} \\
AdaLoRA & \makecell{63.87\% \\ $\pm$7.96\%} & \makecell{\textbf{75.70\%} \\ \textbf{$\pm$2.77\%}} & \makecell{\textbf{77.70\%} \\ \textbf{$\pm$2.81\%}} & \makecell{81.38\% \\ $\pm$1.78\%} & \makecell{82.44\% \\ $\pm$1.56\%} & \makecell{85.66\% \\ $\pm$1.39\%} \\
IA3 & \makecell{66.75\% \\ $\pm$7.65\%} & \makecell{63.56\% \\ $\pm$9.42\%} & \makecell{75.35\% \\ $\pm$3.26\%} & \makecell{\textbf{81.54\%} \\ \textbf{$\pm$0.96\%}} & \makecell{\textbf{82.69\%} \\ \textbf{$\pm$0.97\%}} & \makecell{83.81\% \\ $\pm$1.08\%} \\
\midrule
ICL & \makecell{70.94\% \\ $\pm$3.13\%} & \makecell{70.00\% \\ $\pm$3.27\%} & \makecell{74.06\% \\ $\pm$2.07\%} & \makecell{78.67\% \\ $\pm$1.54\%} & \makecell{81.64\% \\ $\pm$1.11\%} & \makecell{\textbf{88.36\%} \\ \textbf{$\pm$0.45\%}} \\
\bottomrule
\end{tabular}
\caption{Test accuracy (\%) on Web with Llama-3.2-1B for different numbers of training examples $|\mathcal{D}_{train}|$. Best per column in \textbf{bold}.}
\label{tab:cyberbench_web_tc_meta_llama_llama_32_1b}
\end{table}

\begin{table}[h!]
\centering
\begin{tabular}{lcccccc}
\toprule
$|\mathcal{D}_{train}|$ & $10$ & $15$ & $20$ & $30$ & $50$ & $100$ \\
\midrule
AHR (Ours) & \makecell{86.97\% \\ $\pm$1.05\%} & \makecell{87.17\% \\ $\pm$1.10\%} & \makecell{87.91\% \\ $\pm$1.33\%} & \makecell{88.68\% \\ $\pm$1.29\%} & \makecell{89.75\% \\ $\pm$0.96\%} & \makecell{90.66\% \\ $\pm$0.66\%} \\
LoRA & \makecell{86.02\% \\ $\pm$1.83\%} & \makecell{86.29\% \\ $\pm$1.75\%} & \makecell{87.20\% \\ $\pm$0.96\%} & \makecell{83.46\% \\ $\pm$6.29\%} & \makecell{90.56\% \\ $\pm$0.53\%} & \makecell{90.96\% \\ $\pm$0.60\%} \\
AdaLoRA & \makecell{85.84\% \\ $\pm$1.32\%} & \makecell{85.91\% \\ $\pm$2.09\%} & \makecell{\textbf{89.61\%} \\ \textbf{$\pm$1.19\%}} & \makecell{86.89\% \\ $\pm$2.65\%} & \makecell{89.98\% \\ $\pm$1.21\%} & \makecell{\textbf{91.74\%} \\ \textbf{$\pm$0.80\%}} \\
IA3 & \makecell{53.07\% \\ $\pm$13.72\%} & \makecell{81.34\% \\ $\pm$3.68\%} & \makecell{87.57\% \\ $\pm$1.74\%} & \makecell{87.77\% \\ $\pm$1.40\%} & \makecell{\textbf{91.21\%} \\ \textbf{$\pm$0.50\%}} & \makecell{91.15\% \\ $\pm$1.01\%} \\
\midrule
ICL & \makecell{\textbf{87.32\%} \\ \textbf{$\pm$1.04\%}} & \makecell{\textbf{88.99\%} \\ \textbf{$\pm$0.96\%}} & \makecell{88.88\% \\ $\pm$0.87\%} & \makecell{\textbf{90.36\%} \\ \textbf{$\pm$0.38\%}} & \makecell{90.05\% \\ $\pm$0.68\%} & \makecell{91.72\% \\ $\pm$0.27\%} \\
\bottomrule
\end{tabular}
\caption{Test accuracy (\%) on Web with Llama-3.2-3B for different numbers of training examples $|\mathcal{D}_{train}|$. Best per column in \textbf{bold}.}
\label{tab:cyberbench_web_tc_meta_llama_llama_32_3b}
\end{table}

\begin{table}[h!]
\centering
\begin{tabular}{lcccccc}
\toprule
$|\mathcal{D}_{train}|$ & $10$ & $15$ & $20$ & $30$ & $50$ & $100$ \\
\midrule
AHR (Ours) & \makecell{89.07\% \\ $\pm$0.88\%} & \makecell{89.57\% \\ $\pm$0.95\%} & \makecell{88.79\% \\ $\pm$1.18\%} & \makecell{89.34\% \\ $\pm$0.91\%} & \makecell{90.23\% \\ $\pm$0.82\%} & \makecell{90.94\% \\ $\pm$0.42\%} \\
LoRA & \makecell{76.32\% \\ $\pm$8.40\%} & \makecell{78.67\% \\ $\pm$8.64\%} & \makecell{80.97\% \\ $\pm$8.57\%} & \makecell{89.65\% \\ $\pm$0.93\%} & \makecell{\textbf{91.46\%} \\ \textbf{$\pm$0.71\%}} & \makecell{\textbf{92.34\%} \\ \textbf{$\pm$0.39\%}} \\
AdaLoRA & \makecell{86.47\% \\ $\pm$1.96\%} & \makecell{89.86\% \\ $\pm$0.95\%} & \makecell{79.99\% \\ $\pm$5.88\%} & \makecell{82.37\% \\ $\pm$7.82\%} & \makecell{90.82\% \\ $\pm$0.93\%} & \makecell{92.14\% \\ $\pm$0.48\%} \\
IA3 & \makecell{60.45\% \\ $\pm$12.54\%} & \makecell{79.05\% \\ $\pm$8.45\%} & \makecell{89.38\% \\ $\pm$0.71\%} & \makecell{85.11\% \\ $\pm$2.34\%} & \makecell{91.22\% \\ $\pm$0.66\%} & \makecell{92.23\% \\ $\pm$0.59\%} \\
\midrule
ICL & \makecell{\textbf{89.30\%} \\ \textbf{$\pm$0.82\%}} & \makecell{\textbf{90.59\%} \\ \textbf{$\pm$0.63\%}} & \makecell{\textbf{90.56\%} \\ \textbf{$\pm$0.45\%}} & \makecell{\textbf{90.59\%} \\ \textbf{$\pm$0.35\%}} & \makecell{90.38\% \\ $\pm$0.34\%} & \makecell{91.97\% \\ $\pm$0.30\%} \\
\bottomrule
\end{tabular}
\caption{Test accuracy (\%) on Web with Qwen3-8B for different numbers of training examples $|\mathcal{D}_{train}|$. Best per column in \textbf{bold}.}
\label{tab:cyberbench_web_tc_qwen_qwen3_8b}
\end{table}

\begin{table}[h!]
\centering
\begin{tabular}{lcccccc}
\toprule
$|\mathcal{D}_{train}|$ & $10$ & $15$ & $20$ & $30$ & $50$ & $100$ \\
\midrule
AHR (Ours) & \makecell{59.48\% \\ $\pm$1.08\%} & \makecell{58.88\% \\ $\pm$1.56\%} & \makecell{59.22\% \\ $\pm$1.67\%} & \makecell{58.66\% \\ $\pm$2.16\%} & \makecell{61.17\% \\ $\pm$0.54\%} & \makecell{59.33\% \\ $\pm$1.60\%} \\
LoRA & \makecell{51.12\% \\ $\pm$5.78\%} & \makecell{36.94\% \\ $\pm$8.28\%} & \makecell{34.12\% \\ $\pm$9.05\%} & \makecell{50.71\% \\ $\pm$5.72\%} & \makecell{54.64\% \\ $\pm$5.83\%} & \makecell{62.15\% \\ $\pm$1.95\%} \\
AdaLoRA & \makecell{53.51\% \\ $\pm$2.58\%} & \makecell{56.45\% \\ $\pm$2.38\%} & \makecell{58.22\% \\ $\pm$1.73\%} & \makecell{\textbf{58.87\%} \\ \textbf{$\pm$2.09\%}} & \makecell{\textbf{61.82\%} \\ \textbf{$\pm$0.59\%}} & \makecell{\textbf{64.17\%} \\ \textbf{$\pm$1.29\%}} \\
IA3 & \makecell{42.53\% \\ $\pm$7.20\%} & \makecell{48.43\% \\ $\pm$5.83\%} & \makecell{40.67\% \\ $\pm$8.47\%} & \makecell{57.93\% \\ $\pm$2.18\%} & \makecell{60.57\% \\ $\pm$2.23\%} & \makecell{62.16\% \\ $\pm$1.58\%} \\
\midrule
ICL & \makecell{\textbf{61.14\%} \\ \textbf{$\pm$0.11\%}} & \makecell{\textbf{60.93\%} \\ \textbf{$\pm$0.04\%}} & \makecell{\textbf{60.94\%} \\ \textbf{$\pm$0.04\%}} & -- & -- & -- \\
\bottomrule
\end{tabular}
\caption{Test accuracy (\%) on ToxiGen with GPT2-XL for different numbers of training examples $|\mathcal{D}_{train}|$. Best per column in \textbf{bold}.}
\label{tab:toxigen_gpt2_xl}
\end{table}

\begin{table}[h!]
\centering
\begin{tabular}{lcccccc}
\toprule
$|\mathcal{D}_{train}|$ & $10$ & $15$ & $20$ & $30$ & $50$ & $100$ \\
\midrule
AHR (Ours) & \makecell{\textbf{72.37\%} \\ \textbf{$\pm$0.95\%}} & \makecell{71.03\% \\ $\pm$2.05\%} & \makecell{\textbf{72.43\%} \\ \textbf{$\pm$1.36\%}} & \makecell{\textbf{73.16\%} \\ \textbf{$\pm$1.12\%}} & \makecell{\textbf{74.73\%} \\ \textbf{$\pm$0.98\%}} & \makecell{74.47\% \\ $\pm$0.61\%} \\
LoRA & \makecell{67.82\% \\ $\pm$1.85\%} & \makecell{58.31\% \\ $\pm$7.72\%} & \makecell{70.46\% \\ $\pm$1.76\%} & \makecell{67.24\% \\ $\pm$7.14\%} & \makecell{74.37\% \\ $\pm$1.00\%} & \makecell{75.21\% \\ $\pm$0.77\%} \\
AdaLoRA & \makecell{61.84\% \\ $\pm$6.66\%} & \makecell{70.55\% \\ $\pm$1.97\%} & \makecell{70.54\% \\ $\pm$1.59\%} & \makecell{72.14\% \\ $\pm$2.13\%} & \makecell{74.06\% \\ $\pm$1.16\%} & \makecell{\textbf{75.77\%} \\ \textbf{$\pm$0.81\%}} \\
IA3 & \makecell{58.20\% \\ $\pm$8.18\%} & \makecell{60.23\% \\ $\pm$7.27\%} & \makecell{61.86\% \\ $\pm$6.65\%} & \makecell{72.88\% \\ $\pm$1.39\%} & \makecell{72.59\% \\ $\pm$1.15\%} & \makecell{74.03\% \\ $\pm$0.84\%} \\
\midrule
ICL & \makecell{70.40\% \\ $\pm$1.45\%} & \makecell{\textbf{71.09\%} \\ \textbf{$\pm$1.63\%}} & \makecell{70.69\% \\ $\pm$1.47\%} & \makecell{67.62\% \\ $\pm$1.56\%} & \makecell{63.02\% \\ $\pm$0.57\%} & \makecell{65.19\% \\ $\pm$1.23\%} \\
\bottomrule
\end{tabular}
\caption{Test accuracy (\%) on ToxiGen with Llama-3.2-1B for different numbers of training examples $|\mathcal{D}_{train}|$. Best per column in \textbf{bold}.}
\label{tab:toxigen_meta_llama_llama_32_1b}
\end{table}

\begin{table}[h!]
\centering
\begin{tabular}{lcccccc}
\toprule
$|\mathcal{D}_{train}|$ & $10$ & $15$ & $20$ & $30$ & $50$ & $100$ \\
\midrule
AHR (Ours) & \makecell{76.43\% \\ $\pm$1.62\%} & \makecell{77.26\% \\ $\pm$1.55\%} & \makecell{76.44\% \\ $\pm$1.88\%} & \makecell{79.13\% \\ $\pm$1.26\%} & \makecell{81.16\% \\ $\pm$0.65\%} & \makecell{80.49\% \\ $\pm$0.49\%} \\
LoRA & \makecell{71.06\% \\ $\pm$7.59\%} & \makecell{\textbf{79.16\%} \\ \textbf{$\pm$0.65\%}} & \makecell{77.44\% \\ $\pm$1.92\%} & \makecell{79.36\% \\ $\pm$2.06\%} & \makecell{81.43\% \\ $\pm$0.75\%} & \makecell{81.37\% \\ $\pm$0.67\%} \\
AdaLoRA & \makecell{72.45\% \\ $\pm$3.93\%} & \makecell{71.05\% \\ $\pm$5.78\%} & \makecell{77.34\% \\ $\pm$1.95\%} & \makecell{\textbf{80.95\%} \\ \textbf{$\pm$0.79\%}} & \makecell{\textbf{81.55\%} \\ \textbf{$\pm$0.77\%}} & \makecell{\textbf{82.27\%} \\ \textbf{$\pm$0.53\%}} \\
IA3 & \makecell{71.46\% \\ $\pm$4.59\%} & \makecell{72.28\% \\ $\pm$2.95\%} & \makecell{70.15\% \\ $\pm$7.57\%} & \makecell{72.18\% \\ $\pm$7.65\%} & \makecell{81.46\% \\ $\pm$0.66\%} & \makecell{81.40\% \\ $\pm$0.73\%} \\
\midrule
ICL & \makecell{\textbf{77.56\%} \\ \textbf{$\pm$1.82\%}} & \makecell{78.48\% \\ $\pm$0.88\%} & \makecell{\textbf{79.13\%} \\ \textbf{$\pm$1.06\%}} & \makecell{77.43\% \\ $\pm$1.02\%} & \makecell{79.04\% \\ $\pm$0.43\%} & \makecell{79.29\% \\ $\pm$0.47\%} \\
\bottomrule
\end{tabular}
\caption{Test accuracy (\%) on ToxiGen with Llama-3.2-3B for different numbers of training examples $|\mathcal{D}_{train}|$. Best per column in \textbf{bold}.}
\label{tab:toxigen_meta_llama_llama_32_3b}
\end{table}

\begin{table}[h!]
\centering
\begin{tabular}{lcccccc}
\toprule
$|\mathcal{D}_{train}|$ & $10$ & $15$ & $20$ & $30$ & $50$ & $100$ \\
\midrule
AHR (Ours) & \makecell{82.86\% \\ $\pm$0.70\%} & \makecell{82.91\% \\ $\pm$0.73\%} & \makecell{82.76\% \\ $\pm$0.65\%} & \makecell{84.11\% \\ $\pm$0.61\%} & \makecell{84.23\% \\ $\pm$0.24\%} & \makecell{84.48\% \\ $\pm$0.35\%} \\
LoRA & \makecell{37.77\% \\ $\pm$12.16\%} & \makecell{67.99\% \\ $\pm$8.19\%} & \makecell{74.37\% \\ $\pm$7.90\%} & \makecell{83.69\% \\ $\pm$0.95\%} & \makecell{84.88\% \\ $\pm$0.38\%} & \makecell{84.73\% \\ $\pm$0.29\%} \\
AdaLoRA & \makecell{71.56\% \\ $\pm$5.78\%} & \makecell{80.70\% \\ $\pm$2.07\%} & \makecell{81.89\% \\ $\pm$1.12\%} & \makecell{83.79\% \\ $\pm$0.42\%} & \makecell{82.24\% \\ $\pm$1.48\%} & \makecell{84.95\% \\ $\pm$0.31\%} \\
IA3 & \makecell{66.35\% \\ $\pm$8.64\%} & \makecell{75.44\% \\ $\pm$4.14\%} & \makecell{82.76\% \\ $\pm$0.78\%} & \makecell{81.54\% \\ $\pm$2.29\%} & \makecell{82.22\% \\ $\pm$1.17\%} & \makecell{84.63\% \\ $\pm$0.40\%} \\
\midrule
ICL & \makecell{\textbf{83.19\%} \\ \textbf{$\pm$0.64\%}} & \makecell{\textbf{83.32\%} \\ \textbf{$\pm$0.56\%}} & \makecell{\textbf{84.14\%} \\ \textbf{$\pm$0.49\%}} & \makecell{\textbf{84.84\%} \\ \textbf{$\pm$0.36\%}} & \makecell{\textbf{85.48\%} \\ \textbf{$\pm$0.27\%}} & \makecell{\textbf{85.60\%} \\ \textbf{$\pm$0.26\%}} \\
\bottomrule
\end{tabular}
\caption{Test accuracy (\%) on ToxiGen with Qwen3-8B for different numbers of training examples $|\mathcal{D}_{train}|$. Best per column in \textbf{bold}.}
\label{tab:toxigen_qwen_qwen3_8b}
\end{table}

\begin{table}[h!]
\centering
\begin{tabular}{lcccccc}
\toprule
$|\mathcal{D}_{train}|$ & $10$ & $15$ & $20$ & $30$ & $50$ & $100$ \\
\midrule
AHR (Ours) & \makecell{\textbf{65.73\%} \\ \textbf{$\pm$3.82\%}} & \makecell{70.76\% \\ $\pm$4.01\%} & \makecell{70.42\% \\ $\pm$3.94\%} & \makecell{\textbf{84.20\%} \\ \textbf{$\pm$2.01\%}} & \makecell{88.28\% \\ $\pm$1.30\%} & \makecell{91.45\% \\ $\pm$1.37\%} \\
LoRA & \makecell{47.82\% \\ $\pm$10.18\%} & \makecell{71.30\% \\ $\pm$6.00\%} & \makecell{72.48\% \\ $\pm$2.79\%} & \makecell{81.91\% \\ $\pm$2.48\%} & \makecell{\textbf{90.92\%} \\ \textbf{$\pm$1.61\%}} & \makecell{\textbf{92.56\%} \\ \textbf{$\pm$0.56\%}} \\
AdaLoRA & \makecell{56.26\% \\ $\pm$6.62\%} & \makecell{\textbf{74.01\%} \\ \textbf{$\pm$3.51\%}} & \makecell{\textbf{76.49\%} \\ \textbf{$\pm$3.71\%}} & \makecell{74.08\% \\ $\pm$8.84\%} & \makecell{67.02\% \\ $\pm$11.20\%} & \makecell{87.14\% \\ $\pm$5.35\%} \\
IA3 & \makecell{46.03\% \\ $\pm$10.03\%} & \makecell{66.26\% \\ $\pm$4.89\%} & \makecell{54.92\% \\ $\pm$11.76\%} & \makecell{74.73\% \\ $\pm$4.85\%} & \makecell{72.48\% \\ $\pm$9.33\%} & \makecell{75.80\% \\ $\pm$11.19\%} \\
\midrule
ICL & \makecell{57.33\% \\ $\pm$1.97\%} & \makecell{55.80\% \\ $\pm$2.34\%} & -- & -- & -- & -- \\
\bottomrule
\end{tabular}
\caption{Test accuracy (\%) on Jailbreak with GPT2-XL for different numbers of training examples $|\mathcal{D}_{train}|$. Best per column in \textbf{bold}.}
\label{tab:jailbreak_gpt2_xl}
\end{table}

\begin{table}[h!]
\centering
\begin{tabular}{lcccccc}
\toprule
$|\mathcal{D}_{train}|$ & $10$ & $15$ & $20$ & $30$ & $50$ & $100$ \\
\midrule
AHR (Ours) & \makecell{\textbf{79.16\%} \\ \textbf{$\pm$2.25\%}} & \makecell{80.00\% \\ $\pm$3.55\%} & \makecell{81.53\% \\ $\pm$2.03\%} & \makecell{86.41\% \\ $\pm$1.27\%} & \makecell{88.97\% \\ $\pm$0.99\%} & \makecell{90.57\% \\ $\pm$1.12\%} \\
LoRA & \makecell{63.21\% \\ $\pm$10.45\%} & \makecell{66.15\% \\ $\pm$9.96\%} & \makecell{86.45\% \\ $\pm$2.21\%} & \makecell{88.63\% \\ $\pm$1.62\%} & \makecell{90.95\% \\ $\pm$1.19\%} & \makecell{94.43\% \\ $\pm$0.66\%} \\
AdaLoRA & \makecell{71.49\% \\ $\pm$8.25\%} & \makecell{55.46\% \\ $\pm$11.32\%} & \makecell{\textbf{86.87\%} \\ \textbf{$\pm$1.39\%}} & \makecell{62.21\% \\ $\pm$12.91\%} & \makecell{91.79\% \\ $\pm$1.06\%} & \makecell{93.02\% \\ $\pm$1.12\%} \\
IA3 & \makecell{77.52\% \\ $\pm$3.18\%} & \makecell{73.36\% \\ $\pm$8.10\%} & \makecell{81.26\% \\ $\pm$3.28\%} & \makecell{86.60\% \\ $\pm$1.50\%} & \makecell{90.19\% \\ $\pm$0.72\%} & \makecell{87.52\% \\ $\pm$3.70\%} \\
\midrule
ICL & \makecell{74.73\% \\ $\pm$3.09\%} & \makecell{\textbf{81.22\%} \\ \textbf{$\pm$2.96\%}} & \makecell{86.76\% \\ $\pm$2.52\%} & \makecell{\textbf{91.07\%} \\ \textbf{$\pm$1.19\%}} & \makecell{\textbf{95.69\%} \\ \textbf{$\pm$0.33\%}} & \makecell{\textbf{96.95\%} \\ \textbf{$\pm$0.28\%}} \\
\bottomrule
\end{tabular}
\caption{Test accuracy (\%) on Jailbreak with Llama-3.2-1B for different numbers of training examples $|\mathcal{D}_{train}|$. Best per column in \textbf{bold}.}
\label{tab:jailbreak_meta_llama_llama_32_1b}
\end{table}

\begin{table}[h!]
\centering
\begin{tabular}{lcccccc}
\toprule
$|\mathcal{D}_{train}|$ & $10$ & $15$ & $20$ & $30$ & $50$ & $100$ \\
\midrule
AHR (Ours) & \makecell{85.38\% \\ $\pm$3.14\%} & \makecell{87.40\% \\ $\pm$1.56\%} & \makecell{90.23\% \\ $\pm$1.33\%} & \makecell{\textbf{92.48\%} \\ \textbf{$\pm$0.62\%}} & \makecell{92.94\% \\ $\pm$0.64\%} & \makecell{93.85\% \\ $\pm$0.55\%} \\
LoRA & \makecell{84.16\% \\ $\pm$3.55\%} & \makecell{\textbf{90.65\%} \\ \textbf{$\pm$1.26\%}} & \makecell{90.00\% \\ $\pm$1.08\%} & \makecell{76.72\% \\ $\pm$9.76\%} & \makecell{92.60\% \\ $\pm$0.94\%} & \makecell{94.16\% \\ $\pm$0.77\%} \\
AdaLoRA & \makecell{76.45\% \\ $\pm$7.65\%} & \makecell{87.63\% \\ $\pm$1.97\%} & \makecell{76.83\% \\ $\pm$8.66\%} & \makecell{88.74\% \\ $\pm$1.85\%} & \makecell{92.33\% \\ $\pm$0.93\%} & \makecell{84.92\% \\ $\pm$9.00\%} \\
IA3 & \makecell{\textbf{86.56\%} \\ \textbf{$\pm$2.28\%}} & \makecell{84.08\% \\ $\pm$2.12\%} & \makecell{88.28\% \\ $\pm$1.45\%} & \makecell{90.57\% \\ $\pm$0.99\%} & \makecell{92.44\% \\ $\pm$0.87\%} & \makecell{87.94\% \\ $\pm$6.16\%} \\
\midrule
ICL & \makecell{85.46\% \\ $\pm$1.86\%} & \makecell{87.79\% \\ $\pm$1.05\%} & \makecell{\textbf{91.22\%} \\ \textbf{$\pm$0.80\%}} & \makecell{92.10\% \\ $\pm$1.10\%} & \makecell{\textbf{95.46\%} \\ \textbf{$\pm$0.40\%}} & \makecell{\textbf{97.10\%} \\ \textbf{$\pm$0.27\%}} \\
\bottomrule
\end{tabular}
\caption{Test accuracy (\%) on Jailbreak with Llama-3.2-3B for different numbers of training examples $|\mathcal{D}_{train}|$. Best per column in \textbf{bold}.}
\label{tab:jailbreak_meta_llama_llama_32_3b}
\end{table}

\begin{table}[h!]
\centering
\begin{tabular}{lcccccc}
\toprule
$|\mathcal{D}_{train}|$ & $10$ & $15$ & $20$ & $30$ & $50$ & $100$ \\
\midrule
AHR (Ours) & \makecell{88.47\% \\ $\pm$2.61\%} & \makecell{90.46\% \\ $\pm$2.05\%} & \makecell{\textbf{92.33\%} \\ \textbf{$\pm$1.43\%}} & \makecell{\textbf{93.97\%} \\ \textbf{$\pm$0.55\%}} & \makecell{94.43\% \\ $\pm$0.39\%} & \makecell{95.46\% \\ $\pm$0.53\%} \\
LoRA & \makecell{83.17\% \\ $\pm$8.80\%} & \makecell{70.95\% \\ $\pm$11.45\%} & \makecell{79.27\% \\ $\pm$9.45\%} & \makecell{92.40\% \\ $\pm$2.07\%} & \makecell{90.15\% \\ $\pm$3.91\%} & \makecell{\textbf{95.88\%} \\ \textbf{$\pm$0.57\%}} \\
AdaLoRA & \makecell{85.11\% \\ $\pm$3.45\%} & \makecell{\textbf{93.24\%} \\ \textbf{$\pm$0.81\%}} & \makecell{82.21\% \\ $\pm$8.73\%} & \makecell{93.09\% \\ $\pm$0.98\%} & \makecell{\textbf{95.38\%} \\ \textbf{$\pm$0.60\%}} & \makecell{94.20\% \\ $\pm$1.17\%} \\
IA3 & \makecell{81.64\% \\ $\pm$5.08\%} & \makecell{79.73\% \\ $\pm$8.71\%} & \makecell{83.93\% \\ $\pm$4.32\%} & \makecell{86.53\% \\ $\pm$5.19\%} & \makecell{86.91\% \\ $\pm$6.49\%} & \makecell{86.03\% \\ $\pm$5.59\%} \\
\midrule
ICL & \makecell{\textbf{91.87\%} \\ \textbf{$\pm$0.78\%}} & \makecell{91.34\% \\ $\pm$1.01\%} & \makecell{91.30\% \\ $\pm$0.75\%} & \makecell{92.10\% \\ $\pm$0.71\%} & \makecell{92.98\% \\ $\pm$0.29\%} & \makecell{93.70\% \\ $\pm$0.32\%} \\
\bottomrule
\end{tabular}
\caption{Test accuracy (\%) on Jailbreak with Qwen3-8B for different numbers of training examples $|\mathcal{D}_{train}|$. Best per column in \textbf{bold}.}
\label{tab:jailbreak_qwen_qwen3_8b}
\end{table}

\end{document}